%% file: main.tex
\documentclass[11pt,a4paper]{article}

\usepackage[utf8]{inputenc}
\usepackage[T1]{fontenc}
\usepackage{lmodern}

\usepackage[margin=1in]{geometry}

\usepackage{amsmath,amssymb,amsthm}
\usepackage{mathtools}
\usepackage{bm}

\usepackage{graphicx}
\usepackage{booktabs}
\usepackage{multirow}
\usepackage{float}

\usepackage{enumitem}

\usepackage[dvipsnames]{xcolor}
\usepackage{hyperref}
\hypersetup{
  colorlinks  = true,
  linkcolor   = Blue,
  citecolor   = Green,
  urlcolor    = Maroon,
}
\usepackage{url}

\usepackage[numbers,sort&compress]{natbib}

\usepackage[capitalize,nameinlink]{cleveref}

\usepackage{tikz}
\usepackage{pgfplots}
\pgfplotsset{compat=1.18}
\usepgfplotslibrary{fillbetween}
\usetikzlibrary{positioning, arrows.meta, calc, shapes.geometric, fit}
\usepackage{colortbl}

\usepackage{microtype}

\usepackage{algorithm}
\usepackage{algorithmic}

\newcommand{\qos}{\textsc{Qualixar OS}}

\title{\textbf{\qos{}: A Universal Operating System for AI Agent Orchestration}}

\author{
  Varun Pratap Bhardwaj\\
  Independent Researcher, Solution Architect, India\\
  \texttt{varun.pratap.bhardwaj@gmail.com}\\
  \small ORCID: \href{https://orcid.org/0009-0002-8726-4289}{0009-0002-8726-4289}
}

\date{April 2026}

\begin{document}
\maketitle

% ── Abstract ──────────────────────────────────────────────────────────────────
\input{sections/abstract}

% ── Main Sections ─────────────────────────────────────────────────────────────
\input{sections/introduction}
\input{sections/related}
\input{sections/architecture}
\input{sections/forge}
\input{sections/topologies}
\input{sections/routing}
\input{sections/judges}
\input{sections/attribution}
\input{sections/compatibility}
\input{sections/dashboard}
\input{sections/evaluation}
\input{sections/limitations}
\input{sections/conclusion}

% ── Author Biography ──────────────────────────────────────────────────────────
\input{sections/biography}

% ── References ────────────────────────────────────────────────────────────────
\bibliographystyle{plainnat}
\bibliography{references}

\end{document}

%% file: sections/abstract.tex
\begin{abstract}
We present \qos{}, the first application-layer operating system purpose-built for universal AI agent orchestration. Unlike prior work that addresses kernel-level resource scheduling (AIOS) or single-framework pipelines (AutoGen, CrewAI), \qos{} provides a complete runtime for heterogeneous multi-agent systems spanning 10 LLM providers, 8+ agent frameworks, and 7 communication transports.

We contribute: (1)~\textbf{execution semantics for 12 multi-agent topologies} including grid, forest, mesh, and maker patterns; (2)~\textbf{Forge}, an LLM-driven team design engine with historical strategy memory; (3)~\textbf{three-layer model routing} combining Q-learning, five strategies, and Bayesian POMDP; (4)~a \textbf{consensus-based judge pipeline} for multi-criteria quality assurance; (5)~\textbf{four-layer content attribution} with HMAC signing and steganographic watermarks; (6)~\textbf{universal compatibility} via the Claw Bridge supporting MCP and A2A protocols; (7)~a \textbf{24-tab production dashboard} with visual workflow builder and skill marketplace; (8)~\textbf{dynamic multi-provider model discovery} with live catalog queries across 10 providers; (9)~\textbf{Goodhart detection} for judge integrity via cross-model entropy monitoring; (10)~\textbf{empirical drift bounds} with Jensen--Shannon divergence threshold $\Theta{=}0.877$; (11)~\textbf{self-evolution trilemma navigation} addressing the Chen et~al.\ impossibility result; and (12)~\textbf{design-by-contract behavioral invariants} for agent teams.

\qos{} is validated by 2,821 test cases across 217 event types and 8 quality modules. On a custom 20-task evaluation suite, the system achieves 100\% accuracy at a mean cost of \$0.000039 per task. \qos{} is source-available under the Elastic License 2.0 at \url{https://github.com/qualixar/qualixar-os} (DOI: \href{https://doi.org/10.5281/zenodo.19454219}{10.5281/zenodo.19454219}).
\end{abstract}


%% file: sections/introduction.tex
\section{Introduction}
\label{sec:introduction}

The rapid proliferation of large language model (LLM) agents has created a fragmented landscape where developers must choose between incompatible frameworks---AutoGen~\citep{wu2023autogen}, CrewAI~\citep{crewai2024}, MetaGPT~\citep{hong2023metagpt}, LangGraph~\citep{langgraph2024}---each with distinct agent definitions, execution models, and tooling ecosystems. A developer who builds agents in CrewAI cannot run them in AutoGen without rewriting, and neither framework provides cost tracking, quality assurance, or a management dashboard.

We argue that the AI agent ecosystem requires an \emph{operating system}---not another framework. Analogous to how Linux provides a universal runtime for applications regardless of programming language, an agent OS should provide universal orchestration regardless of the agent framework used.

AIOS~\citep{mei2024aios}, published at COLM 2025, introduced the concept of an LLM agent operating system with kernel-level scheduling and context management. We build on this vision but operate at the \emph{application layer}, focusing on orchestration primitives, user experience, and ecosystem compatibility rather than resource scheduling.

\subsection{Contributions}

Industry data underscores the urgency: while 84\% of organizations use AI, only 33\% trust its outputs~(Stack Overflow 2025), and Gartner projects that 40\%+ of agentic AI projects will be cancelled by 2027 due to inadequate governance and quality control.

\qos{} makes seven contributions:

\begin{enumerate}[leftmargin=*]
  \item \textbf{12 Topology Execution Semantics} (\cref{sec:topologies}): A taxonomy of multi-agent execution patterns with formal termination conditions, message-passing protocols, and aggregation strategies---the most comprehensive topology set in any open system.

  \item \textbf{Forge: LLM-Driven Team Design} (\cref{sec:forge}): An automatic team composition engine that translates natural language task descriptions into complete agent teams with role assignments, topology selection, tool attachment, and model allocation.

  \item \textbf{Three-Layer Model Routing with Dynamic Discovery} (\cref{sec:routing,sec:architecture:discovery}): A meta-learning routing architecture where an epsilon-greedy contextual bandit selects the routing \emph{strategy}, the strategy selects the \emph{model}, and a POMDP strategy uses Bayesian belief-state updates for optimal selection under uncertainty. A live model catalog engine queries 10 provider APIs at startup, enabling automatic routing to newly deployed models without configuration changes.

  \item \textbf{Quality Assurance Pipeline} (\cref{sec:judges}): An 8-module evaluation stack comprising consensus-based judging, Goodhart detection for judge integrity via cross-model entropy monitoring, empirical drift bounds with Jensen--Shannon divergence threshold $\Theta = 0.877$~\citep{bhardwaj2026agentassert}, navigation of the Chen et~al.~\citep{chen2025alignment} alignment trilemma with four escape hatches, and design-by-contract behavioral invariants~\citep{meyer1992applying}.

  \item \textbf{Four-Layer Attribution} (\cref{sec:attribution}): A defense-in-depth attribution system designed to survive content transformation, combining visible credits, HMAC signing, steganographic watermarks, and blockchain timestamping.

  \item \textbf{Universal Compatibility} (\cref{sec:compatibility}): The Claw Bridge imports agents from four major formats (OpenClaw, NemoClaw, DeerFlow, GitAgent) while natively supporting both MCP~\citep{mcp2025} and A2A~\citep{a2a2025} protocols.

  \item \textbf{Production Dashboard \& Marketplace} (\cref{sec:dashboard}): A 24-tab browser-based management interface with a visual workflow builder (9 node types, drag-and-drop), a pre-seeded skill marketplace (25 official entries), and real-time WebSocket telemetry.
\end{enumerate}

We term this design philosophy the \emph{Universal Type-C} principle: analogous to how USB Type-C unified charging, data, and video through a single port, \qos{} unifies agent orchestration through a single command protocol that works identically across CLI, MCP, HTTP, WebSocket, and Docker. The 25-command Universal Command Protocol (UCP) ensures that developers interact with \qos{} through the same interface regardless of transport.

%% file: sections/related.tex
\section{Related Work}
\label{sec:related}

\subsection{Multi-Agent Frameworks}

Our analysis draws on systematic study of 40+ open-source agent systems across five tiers of GitHub adoption.

AutoGen~\citep{wu2023autogen} introduced conversational multi-agent programming but supports only sequential and group-chat topologies. CrewAI~\citep{crewai2024} provides role-based agent teams with sequential and hierarchical execution but lacks cost routing or quality assurance. MetaGPT~\citep{hong2023metagpt} encodes Standard Operating Procedures (SOPs) into agent pipelines but is not framework-agnostic. CAMEL~\citep{li2023camel} pioneered role-playing communication but implements only a single two-agent topology. LangGraph~\citep{langgraph2024} offers DAG-based execution with state machines but no automatic team design or dashboard.

\subsection{Agent Operating Systems}

AIOS~\citep{mei2024aios} is the closest prior work, implementing kernel-level agent scheduling, context management, and memory management with support for non-native agents from ReAct, AutoGen, MetaGPT, and Open-Interpreter. AIOS was evaluated on MINT, HumanEval, and SWE-Bench-Lite. AgentOrchestra~\citep{zhang2025} introduced the TEA protocol achieving 89\% on GAIA but lacks a dashboard or marketplace.

\qos{} differentiates from AIOS by operating at the \emph{application layer}: where AIOS manages kernel resources (scheduling, context, storage access), \qos{} manages \emph{orchestration concerns}---topology execution, team design, cost optimization, quality assurance, and user experience. The two systems are complementary rather than competing.

\subsection{Agent Quality \& Security}

AgentAssert~\citep{bhardwaj2026agentassert} introduced behavioral contracts for autonomous agents with formal JSD drift bounds, compliance tracking, and the reliability index $\Theta$. \qos{} ports these formulas directly into its drift monitoring module (\cref{sec:judges:drift}). AgentAssay~\citep{bhardwaj2026agentassay} proposed token-efficient stochastic testing with 3-valued verdicts and adaptive budgets, achieving 78--100\% cost reduction; its evaluation methodology informs the \qos{} judge pipeline design. SkillFortify~\citep{bhardwaj2026skillfortify} established formal security scanning for agent skill ecosystems with 100\% precision across 22 frameworks; \qos{} integrates its verification approach in the marketplace plugin lifecycle. SuperLocalMemory~\citep{bhardwaj2026slmv2,bhardwaj2026slm} provides the 4-layer cognitive memory architecture (working, episodic, semantic, procedural) with information-geometric foundations that \qos{} adapts as SLM-Lite.

\subsection{Cost-Aware Model Routing}

FrugalGPT~\citep{chen2023frugalgpt} demonstrated cost optimization through LLM cascading. RouteLLM~\citep{ong2024routellm} introduced binary routing between strong and weak models. \qos{} extends these to multi-objective optimization (cost, quality, latency) with a three-layer meta-learning architecture (\cref{sec:routing}).

\subsection{Metric Corruption in LLM Evaluation}

The risk of Goodhart's law in LLM evaluation is well established. Skalse et~al.~\citep{skalse2022defining} formalized reward hacking in reinforcement learning, showing that proxy objectives diverge from true objectives under optimization pressure. Gao et~al.~\citep{gao2023scaling} demonstrated scaling laws for reward model overoptimization, where continued RLHF training eventually degrades true performance despite improving proxy scores. Pan et~al.~\citep{pan2022effects} mapped the effects of reward misspecification across model scales.

In the context of LLM-as-judge systems, these findings imply that an agent orchestrator optimizing for judge approval may produce outputs that score well but fail on dimensions not captured by the judge profile. \qos{} addresses this through its Goodhart detection module (\cref{sec:judges:goodhart}), which monitors cross-model entropy, calibration drift, and score inflation to detect when optimization has diverged from genuine quality improvement.

\subsection{Self-Improving Agent Systems}

Chen et~al.~\citep{chen2025alignment} proved that no alignment method can simultaneously achieve strong optimization, perfect value capture, and robust generalization. This impossibility result constrains any system---including \qos{}'s Forge$\to$Judge$\to$RL loop---that claims autonomous self-improvement.

MAST~\citep{cemri2025mast} introduced a comprehensive failure taxonomy for multi-agent LLM systems, identifying 14 failure modes across 7 frameworks, but did not address the trilemma formally. \qos{} takes a different approach: rather than claiming unbounded self-improvement, it explicitly navigates the trilemma by bounding capability gains and preserving safety through architectural firewalls (\cref{sec:judges:trilemma}).

%% file: sections/architecture.tex
\section{System Architecture}
\label{sec:architecture}

\input{figures/architecture}

\qos{} is organized in six layers (\cref{fig:architecture}):

\begin{enumerate}[leftmargin=*]
  \item \textbf{Presentation Layer}: 24-tab React dashboard with Glassmorphism 2.0 design, Zustand state management (1,077 lines), and real-time WebSocket updates with REST polling fallback.

  \item \textbf{Transport Layer}: Seven communication channels---HTTP/REST (Hono), MCP server/client (bidirectional), CLI (Commander.js), Discord, Telegram, Webhook, and Slack---unified behind a channel abstraction.

  \item \textbf{Orchestration Layer}: The 12-step pipeline (\cref{sec:pipeline}) coordinating Forge, Judge, Router, and Cost Tracker with mid-flight steering (pause/resume/redirect/cancel).

  \item \textbf{Execution Layer}: SwarmEngine dispatches agent teams per topology. Agent Registry manages a 5-state lifecycle (idle $\rightarrow$ working $\rightarrow$ paused/error/terminated).

  \item \textbf{Infrastructure Layer}: SLM-Lite cognitive memory, tool registry with 6 categories, MCP consumer, credential vault (AES-256), and the Claw Bridge for framework compatibility.

  \item \textbf{Persistence Layer}: SQLite database with 49 tables, 1 FTS5 virtual table, and 30+ indexes across 17 migration phases, event sourcing for full audit trail, and checkpoint-based task recovery.
\end{enumerate}

\subsection{12-Step Orchestrator Pipeline}
\label{sec:pipeline}

Every task traverses a deterministic 12-step pipeline implemented in \texttt{orchestrator.ts} (923 lines):

\begin{enumerate}[leftmargin=*,noitemsep]
  \item \textbf{Initialize}: Budget check, task registration, steering setup
  \item \textbf{Memory Injection}: SLM-Lite context recall via \texttt{autoInvoke()}
  \item \textbf{Forge Design}: Automatic team composition (\cref{sec:forge})
  \item \textbf{Simulation}: Optional pre-execution simulation (power mode only)
  \item \textbf{Security Validation}: Policy evaluation; blocked tasks cannot proceed
  \item \textbf{Swarm Execution}: Topology-specific agent dispatch (\cref{sec:topologies})
  \item \textbf{Judge Assessment}: Multi-criteria quality evaluation (\cref{sec:judges})
  \item \textbf{Redesign Loop}: On rejection, returns to step 3 (max 5 iterations, 3$\times$ budget cap); after exhaustion, escalates to human review
  \item \textbf{RL Learning}: Composite reward recording for strategy improvement
  \item \textbf{Behavior Capture}: Per-agent behavioral pattern storage
  \item \textbf{Output Formatting}: Result assembly and disk persistence
  \item \textbf{Finalize}: Database update, event emission, checkpoint cleanup
\end{enumerate}

The orchestrator checks steering state between every major step, enabling mid-flight task control. Paused tasks poll at 100ms intervals with a 1-hour timeout. Redirected tasks restart the pipeline with a new prompt while preserving the task identifier.

\input{figures/task-flow}

The end-to-end task lifecycle is visualized in \cref{fig:taskflow}. As a concrete example: given the prompt ``Build a REST API for user management,'' the orchestrator classifies the task as \texttt{code}, Forge selects a 3-agent pipeline topology (architect, implementer, reviewer), the judge panel evaluates the output against code-specific criteria, and on rejection, Forge redesigns with an alternative debate topology. Quality monitors---Goodhart detection, distributional drift, trilemma bounds, and behavioral contracts---run in parallel during swarm execution and feed guard verdicts into the judge assessment, creating a defense-in-depth evaluation stack.

\subsection{Dual Operating Modes}

\qos{} operates in two modes governed by a feature-gate engine (\texttt{mode-engine.ts}, 200 lines):

\begin{table}[H]
\centering
\caption{Feature gates by operating mode.}
\label{tab:modes}
\begin{tabular}{lcc}
\toprule
\textbf{Feature} & \textbf{Companion} & \textbf{Power} \\
\midrule
Topologies & 6 & All 12 \\
Max judges & 2 & 5 \\
Routing strategies & 3 & 5 (+POMDP, balanced) \\
Reinforcement learning & No & Yes \\
Container isolation & No & Yes \\
Simulation & No & Yes \\
\bottomrule
\end{tabular}
\end{table}

\subsection{Model Discovery \& Dynamic Routing}
\label{sec:architecture:discovery}

\input{figures/discovery-flow}

Rather than relying on static model configuration files, \qos{} discovers available models at runtime by querying provider catalog APIs. The discovery engine (\texttt{model-discovery.ts}, 380 lines) supports 10 providers:

\begin{table}[H]
\centering
\caption{Model discovery: supported providers and their catalog APIs.}
\label{tab:discovery}
\small
\begin{tabular}{@{}lll@{}}
\toprule
\textbf{Provider} & \textbf{Discovery API} & \textbf{Auth} \\
\midrule
Azure AI Foundry & \texttt{/models?api-version=...} & Azure AD \\
OpenAI & \texttt{/v1/models} & API key \\
Anthropic & \texttt{/v1/models} & API key \\
Google (Vertex) & \texttt{/v1/models} & OAuth2 \\
Bedrock & \texttt{ListFoundationModels} & AWS IAM \\
Ollama & \texttt{/api/tags} & Local \\
LM Studio & \texttt{/v1/models} & Local \\
llama.cpp & \texttt{/v1/models} & Local \\
vLLM & \texttt{/v1/models} & Local \\
HuggingFace TGI & \texttt{/info} & API key \\
\bottomrule
\end{tabular}
\end{table}

The full discovery-to-routing flow is illustrated in \cref{fig:discovery}. Discovery runs at system startup and can be triggered on-demand via the dashboard or API. The engine caches results with a configurable TTL (default: 1 hour) and merges discovered models with any static configuration, giving static entries priority for overrides.

Three routing strategies consume discovery results:

\begin{itemize}[leftmargin=*,noitemsep]
  \item \textbf{Quality-first}: Selects the highest-rated model from the discovered catalog, regardless of cost.
  \item \textbf{Balanced}: Weighted combination of quality score and inverse cost, selecting Pareto-optimal models.
  \item \textbf{Cost-first}: Selects the cheapest model that meets a minimum quality threshold.
\end{itemize}

\subsection{Protocol-Unified Agent Teams}
\label{sec:protocol-unified}
A persistent limitation of existing frameworks is the split between internal agent communication (proprietary, in-process) and external communication (standardized protocols like A2A). \qos{} resolves this by adopting A2A as the canonical message format for \emph{all} agents---local and remote. The \texttt{ProtocolRouter} selects the optimal transport (in-memory for co-located agents, HTTP for remote, MCP for tools) while maintaining format consistency. This enables hot-swapping any local agent with a remote service with zero code changes.

\noindent\textbf{Verification.} Live discovery against the Azure AI Foundry (enterprise Azure subscription) returned 236 models, including GPT-5.4-mini, DeepSeek-V3.2-Speciale, Grok-4.1-fast-reasoning, and Claude~Opus~4.6. A round-trip ``Hello'' call through the discovered GPT-5.4-mini endpoint confirmed end-to-end functionality.

%% file: figures/architecture.tex
% Figure 1: Qualixar OS — Full Component Architecture
\begin{figure*}[t]
\centering
\begin{tikzpicture}[
  % Styles
  comp/.style={draw=#1!60!black, fill=#1!12, thick, rounded corners=3pt,
    minimum height=0.75cm, font=\sffamily\scriptsize, text=black, align=center},
  group/.style={draw=#1!40!black, fill=#1!5, thick, rounded corners=5pt,
    inner sep=6pt},
  groupnew/.style={draw=#1!40!black, fill=#1!5, thick, rounded corners=5pt,
    inner sep=6pt, dashed},
  glabel/.style={font=\sffamily\footnotesize\bfseries, color=#1!70!black},
  arr/.style={->, >=stealth, thick, color=gray!60!black},
  dasharr/.style={->, >=stealth, thick, dashed, color=gray!50!black},
  redarr/.style={->, >=stealth, very thick, color=red!70!black},
  newtag/.style={font=\sffamily\tiny\bfseries, color=white,
    fill=ForestGreen!70!black, rounded corners=2pt, inner sep=2pt},
  edgelabel/.style={fill=white, inner sep=1pt, font=\sffamily\tiny},
  background rectangle/.style={fill=white},
]

% ═══════════════════════════════════════════════════════════════════════
% LEFT COLUMN: TRANSPORT / ACCESS (x = -5.8)
% ═══════════════════════════════════════════════════════════════════════
\node[comp=RedOrange, minimum width=2.6cm] (t1) at (-5.8, 0.0) {HTTP/REST (Hono)};
\node[comp=RedOrange, minimum width=2.6cm] (t2) at (-5.8,-0.95) {MCP Server+Client};
\node[comp=RedOrange, minimum width=2.6cm] (t3) at (-5.8,-1.9) {CLI (Commander.js)};
\node[comp=RedOrange, minimum width=2.6cm] (t4) at (-5.8,-2.85) {WebSocket (JSON-RPC)};
\node[comp=RedOrange, minimum width=2.6cm] (t5) at (-5.8,-3.8) {Discord / Telegram};
\node[comp=RedOrange, minimum width=2.6cm] (t6) at (-5.8,-4.75) {Webhook / Slack};

% Transport group label
\node[glabel=RedOrange, anchor=south] at (-5.8, 0.55) {Transport Layer};
\draw[RedOrange!40!black, thick, rounded corners=5pt]
  (-7.4, 0.5) rectangle (-4.2, -5.2);

% ═══════════════════════════════════════════════════════════════════════
% CENTER: CORE ENGINE (x = 0, y = 0...-5)
% ═══════════════════════════════════════════════════════════════════════

% Orchestrator outer box
\draw[YellowOrange!50!black, very thick, rounded corners=6pt, fill=YellowOrange!4]
  (-2.0, 1.2) rectangle (4.8, -5.0);
\node[glabel=YellowOrange, anchor=south] at (1.4, 1.2)
  {\large Orchestrator (12-Step Pipeline)};

% Forge
\node[comp=Emerald, minimum width=2.2cm, minimum height=1.1cm] (forge) at (-0.5, 0.0)
  {{\bfseries Forge}\\Team Design\\Task Classify};

% Swarm Engine
\node[comp=RoyalBlue, minimum width=2.2cm, minimum height=1.1cm] (swarm) at (-0.5, -1.7)
  {{\bfseries Swarm Engine}\\12 Topologies\\Agent Registry};

% Judge Pipeline
\node[comp=Plum, minimum width=2.2cm, minimum height=1.1cm] (judge) at (2.8, 0.0)
  {{\bfseries Judge Pipeline}\\Multi-criteria\\3 Consensus Algos};

% RL Trainer
\node[comp=ForestGreen, minimum width=2.2cm, minimum height=1.1cm] (rl) at (2.8, -1.7)
  {{\bfseries RL Trainer}\\Q-Learning\\Reward Signals};
\node[font=\sffamily\tiny\bfseries, color=RoyalBlue!70!black,
  fill=RoyalBlue!10, rounded corners=2pt, inner sep=2pt] at (4.35, -1.2) {Pivot~2};

% Model Router
\node[comp=Cyan, minimum width=2.2cm, minimum height=1.1cm] (router) at (-0.5, -3.4)
  {{\bfseries Model Router}\\5 Strategies\\+Discovery};

% Cost Tracker
\node[comp=YellowOrange, minimum width=2.2cm, minimum height=0.85cm] (cost) at (2.8, -3.4)
  {{\bfseries Cost Tracker}\\Budget + Attribution};

% --- Internal arrows ---
\draw[arr] (forge) -- (swarm) node[midway, left, edgelabel] {dispatch};
\draw[arr] (swarm.east) -- ++(0.6,0) |- (judge.south west)
  node[near end, below, edgelabel] {results};
\draw[arr] (judge) -- (rl) node[midway, right, edgelabel] {reward};
\draw[arr] (rl.south) -- ++(0,-0.3) -| (router.east)
  node[near start, right, edgelabel] {update};
\draw[arr] (router) -- (swarm) node[midway, left, edgelabel, font=\sffamily\small] {model};

% Reject -> Redesign loop (RED)
\draw[redarr, rounded corners=4pt]
  (judge.north) -- ++(0, 0.45) -- ++(-3.3, 0) -- (forge.north)
  node[pos=0.5, above=2pt, fill=white, inner sep=2pt, font=\sffamily\tiny\bfseries, text=red!70!black]
  {reject $\to$ redesign (max 5)};

% ═══════════════════════════════════════════════════════════════════════
% RIGHT COLUMN: QUALITY GUARDS (x = 6.2)
% ═══════════════════════════════════════════════════════════════════════
\node[comp=ForestGreen, minimum width=2.6cm] (q1) at (6.6, 0.0)
  {Goodhart Detector};
\node[comp=ForestGreen, minimum width=2.6cm] (q2) at (6.6, -0.95)
  {Drift Monitor\\{\tiny Distribution Drift}};
\node[comp=ForestGreen, minimum width=2.6cm] (q3) at (6.6, -1.9)
  {Trilemma Guard\\{\tiny 4 escape hatches}};
\node[comp=ForestGreen, minimum width=2.6cm] (q4) at (6.6, -2.85)
  {Behavioral Contracts\\{\tiny pre/post verify}};
\node[comp=ForestGreen, minimum width=2.6cm] (q5) at (6.6, -3.8)
  {Forge Memory Guard};

% Quality Guards group
\node[glabel=ForestGreen, anchor=south] at (6.6, 0.55) {Quality Guards};
\node[newtag] at (8.3, 0.55) {NEW};
\draw[ForestGreen!40!black, thick, rounded corners=5pt, dashed]
  (4.9, 0.5) rectangle (8.3, -4.25);

% Quality arrows (guards chain)
\draw[arr, color=ForestGreen!50!black] (q1) -- (q2);
\draw[arr, color=ForestGreen!50!black] (q2) -- (q3);
\draw[arr, color=ForestGreen!50!black] (q3) -- (q4);
\draw[arr, color=ForestGreen!50!black] (q4) -- (q5);

% ═══════════════════════════════════════════════════════════════════════
% FAR RIGHT: MONITORING (x = 9.6)
% ═══════════════════════════════════════════════════════════════════════
\node[comp=Cyan, minimum width=2.0cm, minimum height=1.0cm] (ebus) at (9.8, -1.0)
  {{\bfseries EventBus}\\{\tiny 217 event types}};
\node[comp=Cyan, minimum width=2.0cm, minimum height=1.0cm] (dash) at (9.8, -2.8)
  {{\bfseries Dashboard}\\{\tiny 24 tabs, WS}};

\draw[arr] (ebus) -- (dash);
\node[glabel=Cyan, anchor=south] at (9.8, -0.15) {Monitoring};
\draw[Cyan!40!black, thick, rounded corners=5pt]
  (8.5, -0.2) rectangle (11.1, -3.5);

% ═══════════════════════════════════════════════════════════════════════
% BOTTOM: INFRASTRUCTURE (y = -6.0)
% ═══════════════════════════════════════════════════════════════════════
\node[comp=Plum, minimum width=2.0cm] (slm) at (-4.5, -6.4)
  {SLM-Lite\\{\tiny 4-layer memory}};
\node[comp=Plum, minimum width=2.0cm] (sqlite) at (-2.0, -6.4)
  {SQLite\\{\tiny 49 tables, ES}};
\node[comp=Plum, minimum width=2.0cm] (tools) at (0.5, -6.4)
  {Tool Registry\\{\tiny 6 categories}};
\node[comp=Plum, minimum width=2.0cm] (vault) at (3.0, -6.4)
  {Credential Vault\\{\tiny AES-256}};
\node[comp=Plum, minimum width=1.8cm] (bridge) at (5.3, -6.4)
  {Claw Bridge\\{\tiny 4 parsers}};

% Infrastructure group
\node[glabel=Plum, anchor=north] at (0.5, -7.25) {Infrastructure \& Persistence};
\draw[Plum!40!black, thick, rounded corners=5pt]
  (-5.8, -5.8) rectangle (6.5, -7.15);

% Model Discovery (between router and infrastructure)
\node[comp=RoyalBlue, minimum width=2.4cm, font=\sffamily\tiny] (disc) at (-0.5, -5.2)
  {Model Discovery\\10 Providers};
\node[newtag] at (0.9, -5.2) {NEW};

% ═══════════════════════════════════════════════════════════════════════
% CROSS-SYSTEM ARROWS
% ═══════════════════════════════════════════════════════════════════════

% Transport -> Orchestrator
\draw[arr, very thick] (-4.2, -2.4) -- (-2.0, -2.4)
  node[midway, above, edgelabel] {unified channel};

% Judge -> Quality Guards
\draw[dasharr, color=Plum!60!black]
  (judge.east) -- ++(0.5,0) |- (q1.west)
  node[near start, above, edgelabel] {assess};

% Quality Guards -> EventBus
\draw[dasharr, color=ForestGreen!60!black]
  (q3.east) -| (ebus.south west)
  node[near start, above, fill=white, inner sep=2pt, font=\sffamily\tiny, text=ForestGreen!60!black] {events};

% Orchestrator -> EventBus
\draw[dasharr, color=YellowOrange!60!black]
  (4.8, -1.0) -- (ebus.west)
  node[midway, above, edgelabel] {lifecycle};

% Router -> Model Discovery
\draw[arr] (router.south) -- (disc.north);

% Orchestrator -> Infrastructure (persist)
\draw[arr] (1.4, -5.0) -- (1.4, -5.8)
  node[midway, right, edgelabel] {persist};

% SLM -> Forge (memory injection)
\draw[dasharr, color=Plum!60!black]
  (slm.north) -- ++(0, 1.4) -| (-2.0, 0.0) -- (forge.west)
  node[near start, left, fill=white, inner sep=2pt, font=\sffamily\tiny, text=Plum!60!black] {context recall};

\end{tikzpicture}
\caption{Full component architecture of \qos{}. The core engine (center, yellow border) houses the Orchestrator's 12-step pipeline with Forge, Swarm, Judge, Router, RL Trainer, and Cost Tracker. Seven transport channels (left) provide universal access. Quality guards (right, dashed green border) are new in Pivot~2 and emit events to the central EventBus/Dashboard monitoring stack (far right). Infrastructure spans the bottom. Red arrow: the reject$\to$redesign feedback loop. Dashed arrows: event and feedback flows.}
\label{fig:architecture}
\end{figure*}
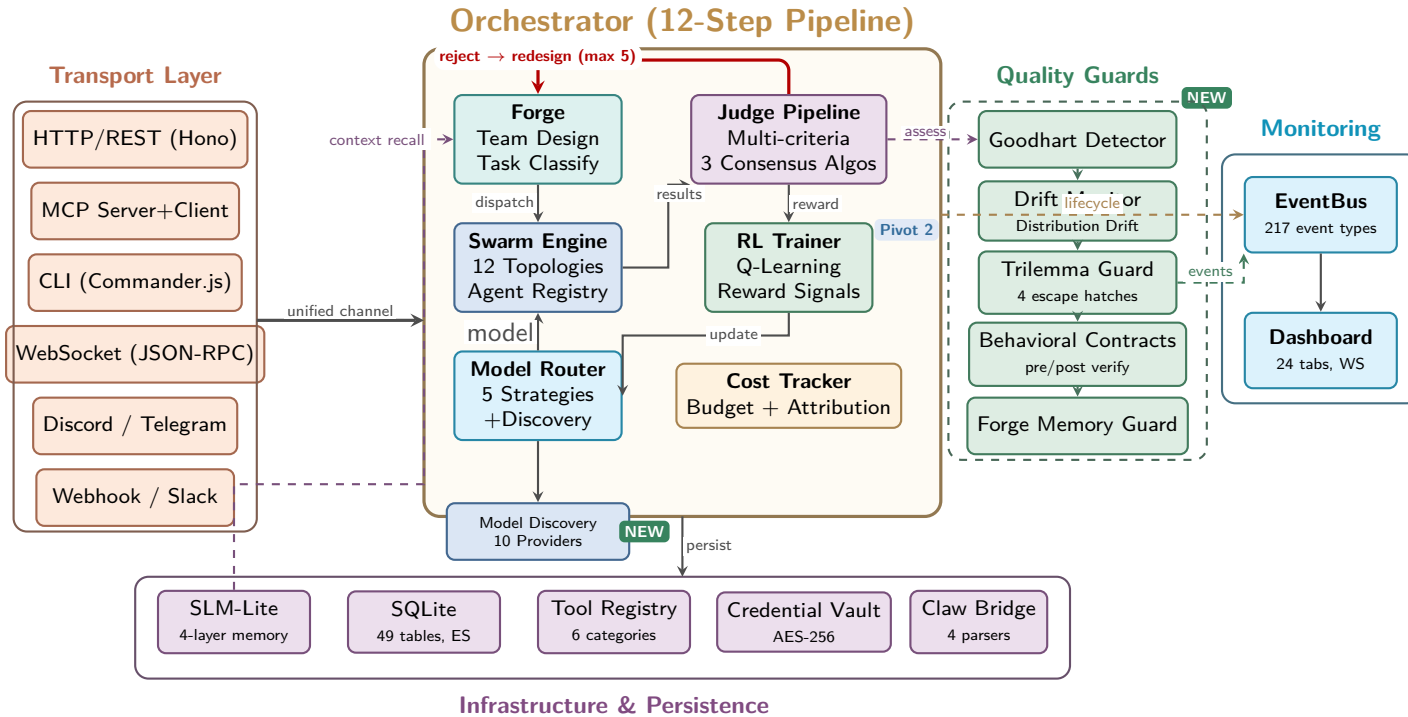

%% file: figures/task-flow.tex
% Figure 2: End-to-End Task Lifecycle Flow
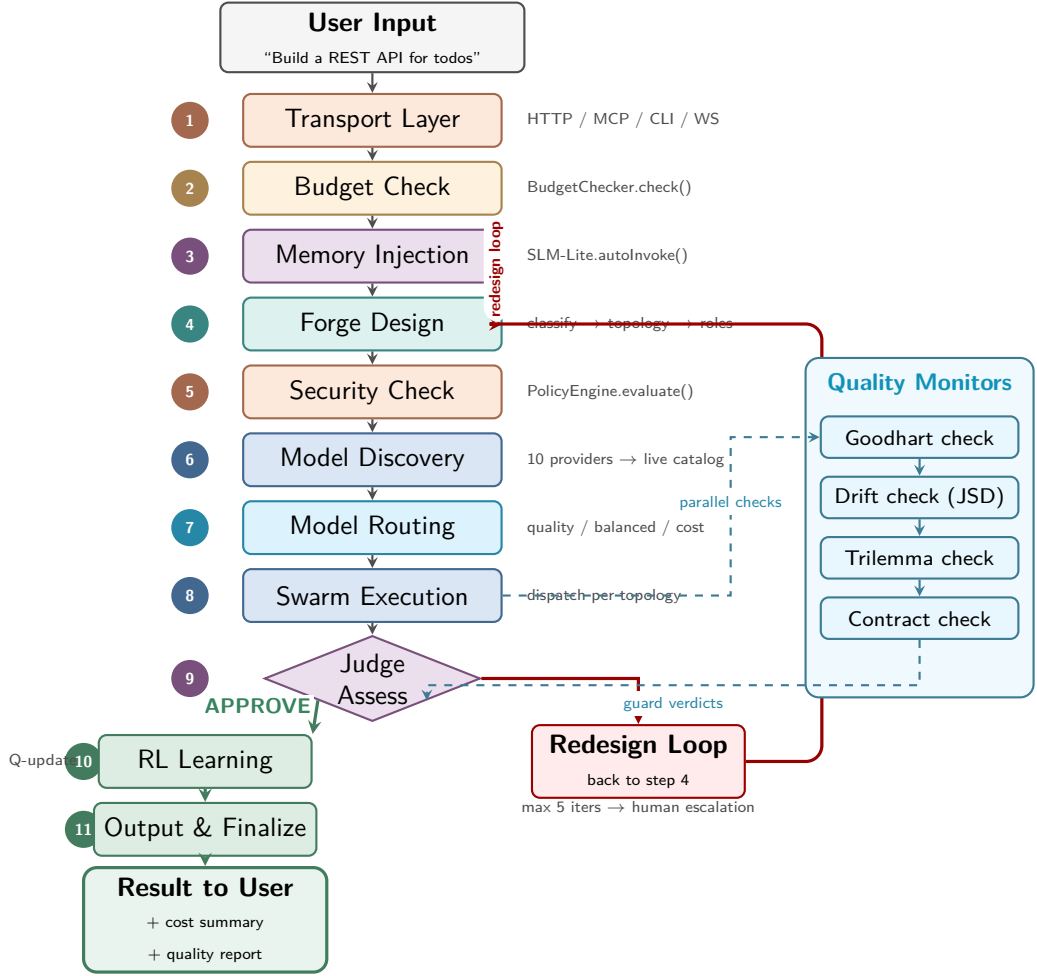
\begin{figure*}[t]
\centering
\begin{tikzpicture}[
  % Step box
  step/.style={draw=#1!60!black, fill=#1!12, thick, rounded corners=3pt,
    minimum width=3.4cm, minimum height=0.7cm, font=\sffamily\small,
    text=black, align=center},
  % Decision diamond
  decision/.style={draw=Plum!60!black, fill=Plum!12, thick,
    diamond, aspect=2.5, inner sep=1pt, font=\sffamily\small,
    text=black, align=center, minimum width=2cm},
  % Small annotation
  annot/.style={font=\sffamily\tiny, color=gray!60!black},
  % Edge label with white background
  edgelabel/.style={fill=white, inner sep=1pt, font=\sffamily\tiny},
  % Number circle
  stepnum/.style={circle, fill=#1!60!black, text=white,
    font=\sffamily\tiny\bfseries, inner sep=1.5pt, minimum size=14pt},
  % Arrows
  arr/.style={->, >=stealth, thick, color=gray!60!black},
  garr/.style={->, >=stealth, very thick, color=ForestGreen!60!black},
  rarr/.style={->, >=stealth, very thick, color=red!60!black},
  % Dashed arrow
  dasharr/.style={->, >=stealth, thick, dashed, color=gray!50!black},
  % Monitor sidebar box
  monitor/.style={draw=Cyan!50!black, fill=Cyan!8, thick, rounded corners=3pt,
    minimum width=2.6cm, minimum height=0.55cm, font=\sffamily\scriptsize,
    text=black, align=center},
]

% ═══════════════════════════════════════════════════════════════════════
% MAIN FLOW (vertical, x=0)
% ═══════════════════════════════════════════════════════════════════════

% User input
\node[step=gray, minimum width=4cm, fill=gray!8] (user) at (0, 0)
  {{\bfseries User Input}\\{\tiny ``Build a REST API for todos''}};

% Step 1: Transport
\node[stepnum=RedOrange] (n1) at (-2.4, -1.1) {1};
\node[step=RedOrange] (s1) at (0, -1.1) {Transport Layer};
\node[annot, right] at (1.9, -1.1) {HTTP / MCP / CLI / WS};

% Step 2: Budget Check
\node[stepnum=YellowOrange] (n2) at (-2.4, -2.0) {2};
\node[step=YellowOrange] (s2) at (0, -2.0) {Budget Check};
\node[annot, right] at (1.9, -2.0) {BudgetChecker.check()};

% Step 3: Memory Injection
\node[stepnum=Plum] (n3) at (-2.4, -2.9) {3};
\node[step=Plum] (s3) at (0, -2.9) {Memory Injection};
\node[annot, right] at (1.9, -2.9) {SLM-Lite.autoInvoke()};

% Step 4: Forge Design
\node[stepnum=Emerald] (n4) at (-2.4, -3.8) {4};
\node[step=Emerald] (s4) at (0, -3.8) {Forge Design};
\node[annot, right, text width=3.5cm] at (1.9, -3.8) {classify $\to$ topology $\to$ roles};

% Step 5: Security
\node[stepnum=RedOrange] (n5) at (-2.4, -4.7) {5};
\node[step=RedOrange] (s5) at (0, -4.7) {Security Check};
\node[annot, right] at (1.9, -4.7) {PolicyEngine.evaluate()};

% Step 6: Model Discovery
\node[stepnum=RoyalBlue] (n6) at (-2.4, -5.6) {6};
\node[step=RoyalBlue] (s6) at (0, -5.6) {Model Discovery};
\node[annot, right] at (1.9, -5.6) {10 providers $\to$ live catalog};

% Step 7: Model Routing
\node[stepnum=Cyan] (n7) at (-2.4, -6.5) {7};
\node[step=Cyan] (s7) at (0, -6.5) {Model Routing};
\node[annot, right] at (1.9, -6.5) {quality / balanced / cost};

% Step 8: Swarm Execution
\node[stepnum=RoyalBlue] (n8) at (-2.4, -7.4) {8};
\node[step=RoyalBlue] (s8) at (0, -7.4) {Swarm Execution};
\node[annot, right] at (1.9, -7.4) {dispatch per topology}; %chktex 8

% Step 9: Judge Assessment (decision point)
\node[stepnum=Plum] (n9) at (-2.4, -8.5) {9};
\node[decision] (s9) at (0, -8.5) {Judge\\Assess};

% ═══════════════════════════════════════════════════════════════════════
% DECISION BRANCHES
% ═══════════════════════════════════════════════════════════════════════

% APPROVE (green, goes down-left)
\node[stepnum=ForestGreen] (n10) at (-3.8, -9.6) {10};
\node[step=ForestGreen, minimum width=2.8cm] (s10) at (-2.2, -9.6)
  {RL Learning};
\node[annot, left, text width=1.5cm, align=right] at (-3.75, -9.6) {Q{-}update};

\node[stepnum=ForestGreen] (n11) at (-3.8, -10.5) {11};
\node[step=ForestGreen, minimum width=2.8cm] (s11) at (-2.2, -10.5)
  {Output \& Finalize};

% Result box
\node[draw=ForestGreen!60!black, fill=ForestGreen!8, very thick,
  rounded corners=5pt, minimum width=3.2cm, minimum height=1.0cm,
  font=\sffamily\small, align=center] (result) at (-2.2, -11.7)
  {{\bfseries Result to User}\\{\tiny + cost summary}\\{\tiny + quality report}};

% REJECT (red, goes right and loops back)
\node[step=red, minimum width=2.8cm, fill=red!8, draw=red!60!black] (reject) at (3.5, -9.6)
  {{\bfseries Redesign Loop}\\{\tiny back to step 4}};
\node[annot] at (3.5, -10.2) {max 5 iters $\to$ human escalation};

% Decision arrows
\draw[garr] (s9.south west) -- (s10.north east)
  node[midway, above left, fill=white, inner sep=1pt, font=\sffamily\scriptsize\bfseries,
    text=ForestGreen!70!black] {APPROVE};
\draw[rarr] (s9.east) -| (reject.north)
  node[near end, right, fill=white, inner sep=1pt, font=\sffamily\scriptsize\bfseries,
    text=red!70!black] {REJECT};

% Reject loop back to Forge (step 4)
\draw[rarr, rounded corners=6pt]
  (reject.east) -- ++(1.0, 0) -- ++(0, 5.8) -- ++(-4.5, 0) -- (s4.east)
  node[pos=0.25, right=4pt, fill=white, inner sep=2pt, font=\sffamily\tiny\bfseries, text=red!60!black, rotate=90]
  {redesign loop};

% ═══════════════════════════════════════════════════════════════════════
% MAIN FLOW ARROWS (green success path)
% ═══════════════════════════════════════════════════════════════════════
\draw[arr] (user) -- (s1);
\draw[arr] (s1) -- (s2);
\draw[arr] (s2) -- (s3);
\draw[arr] (s3) -- (s4);
\draw[arr] (s4) -- (s5);
\draw[arr] (s5) -- (s6);
\draw[arr] (s6) -- (s7);
\draw[arr] (s7) -- (s8);
\draw[arr] (s8) -- (s9);
\draw[garr] (s10) -- (s11);
\draw[garr] (s11) -- (result);

% ═══════════════════════════════════════════════════════════════════════
% QUALITY MONITORS SIDEBAR (blue, right side)
% ═══════════════════════════════════════════════════════════════════════
\node[draw=Cyan!50!black, fill=Cyan!5, thick, rounded corners=5pt,
  minimum width=3.0cm, minimum height=4.5cm] (monbox) at (7.2, -6.5) {};
\node[font=\sffamily\footnotesize\bfseries, color=Cyan!70!black]
  at (7.2, -4.6) {Quality Monitors};

\node[monitor] (m1) at (7.2, -5.3) {Goodhart check};
\node[monitor] (m2) at (7.2, -6.1) {Drift check (JSD)};
\node[monitor] (m3) at (7.2, -6.9) {Trilemma check};
\node[monitor] (m4) at (7.2, -7.7) {Contract check};

\draw[arr, color=Cyan!50!black] (m1) -- (m2);
\draw[arr, color=Cyan!50!black] (m2) -- (m3);
\draw[arr, color=Cyan!50!black] (m3) -- (m4);

% Connect monitors to judge
\draw[dasharr, color=Cyan!50!black]
  (s8.east) -- ++(3.0, 0) |- (m1.west)
  node[near start, above, fill=white, inner sep=2pt, font=\sffamily\tiny, text=Cyan!60!black, draw=none] {parallel checks};

% Monitor verdict feeds into judge
\draw[dasharr, color=Cyan!50!black]
  (m4.south) -- ++(0, -0.6) -| (s9.south east)
  node[near start, below=2pt, fill=white, inner sep=2pt, font=\sffamily\tiny, text=Cyan!60!black, draw=none] {guard verdicts};

\end{tikzpicture}
\caption{End-to-end task lifecycle in \qos{}. Numbered steps 1--11 trace the 12-step pipeline from user input through transport, memory injection, Forge team design, model discovery and routing, swarm execution, and judge assessment. The diamond decision point routes to either RL learning and output (green path) or redesign (red loop, max 5 iterations). Quality monitors (blue sidebar) run in parallel during execution and feed guard verdicts into the judge assessment.}
\label{fig:taskflow}
\end{figure*}

%% file: figures/discovery-flow.tex
% Figure 3: Model Discovery & Routing Architecture (Enhanced)
\begin{figure}[t]
\centering
\begin{tikzpicture}[
  box/.style={draw=#1!60!black, fill=#1!12, thick, rounded corners=4pt,
    minimum height=0.8cm, font=\sffamily\small, text=black, align=center},
  provider/.style={draw=RoyalBlue!50!black, fill=RoyalBlue!8, thick,
    rounded corners=3pt, minimum height=0.65cm, minimum width=1.7cm,
    font=\sffamily\tiny, text=black, align=center},
  strat/.style={draw=#1!60!black, fill=#1!12, thick, rounded corners=4pt,
    minimum height=0.8cm, minimum width=2.2cm, font=\sffamily\scriptsize,
    text=black, align=center},
  arr/.style={->, >=stealth, thick, color=gray!60!black},
  dasharr/.style={->, >=stealth, thick, dashed, color=Plum!50!black},
  annot/.style={font=\sffamily\tiny, color=gray!60!black},
  edgelabel/.style={fill=white, inner sep=1pt, font=\sffamily\tiny},
]

% ── Config Block ────────────────────────────────────────────────────
\node[box=gray, minimum width=6.5cm, minimum height=1.2cm] (config) at (0, 0) {};
\node[font=\sffamily\scriptsize\bfseries, anchor=north west] at (-3.0, 0.45) {config.yaml};
\node[font=\ttfamily\tiny, anchor=north west, text=gray!70!black] at (-2.8, 0.15) {%
providers:};
\node[font=\ttfamily\tiny, anchor=north west, text=gray!70!black] at (-2.4, -0.1) {%
azure{:} \{endpoint, api\_key\_env\}};
\node[font=\ttfamily\tiny, anchor=north west, text=gray!70!black] at (-2.4, -0.35) {%
ollama: \{endpoint{:} localhost{:}11434\}};
\node[font=\ttfamily\tiny, anchor=north west, text=RoyalBlue!70!black] at (1.2, 0.15) {%
routing{:} balanced};
\node[font=\ttfamily\tiny, anchor=north west, text=RoyalBlue!70!black] at (1.2, -0.1) {%
cache\_ttl{:} 3600};

% ── Discovery Engine ────────────────────────────────────────────────
\draw[Emerald!40!black, very thick, rounded corners=5pt, fill=Emerald!4]
  (-3.5, -1.2) rectangle (3.5, -4.6);
\node[font=\sffamily\footnotesize\bfseries, color=Emerald!70!black, anchor=north]
  at (0, -1.25) {Model Discovery Engine};

% Provider boxes (row)
\node[provider] (p1) at (-2.2, -2.1) {Azure\\{\tiny GET /models}};
\node[provider] (p2) at (-0.2, -2.1) {OpenAI\\{\tiny GET /v1/models}};
\node[provider] (p3) at (1.8, -2.1) {Ollama\\{\tiny GET /api/tags}};
\node[annot] at (3.2, -2.1) {$\times 10$};

% Merge arrow
\draw[arr] (p1.south) -- ++(0, -0.4);
\draw[arr] (p2.south) -- ++(0, -0.4);
\draw[arr] (p3.south) -- ++(0, -0.4);
\draw[arr, very thick] (-2.2, -2.9) -- (1.8, -2.9);

% Live Catalog
\node[box=YellowOrange, minimum width=4.5cm, minimum height=1.0cm] (catalog) at (0, -3.7)
  {{\bfseries Live Catalog}\\{\tiny 236 models $\cdot$ quality scores $\cdot$ pricing $\cdot$ context windows}};

% ── Config -> Engine ────────────────────────────────────────────────
\draw[arr, very thick] (config.south) -- ++(0, -0.4)
  node[midway, right, fill=white, inner sep=2pt, font=\sffamily\tiny, text=gray!60!black] {provider list};

% ── Strategy Selection ──────────────────────────────────────────────
\node[strat=Emerald] (sq) at (-2.8, -5.6) {{\bfseries Quality}\\best model};
\node[strat=RoyalBlue] (sb) at (0, -5.6) {{\bfseries Balanced}\\quality/cost};
\node[strat=RedOrange] (sc) at (2.8, -5.6) {{\bfseries Cost}\\cheapest};

\draw[arr] (catalog.south) -- ++(0, -0.3) -| (sq.north);
\draw[arr] (catalog.south) -- ++(0, -0.3) -- (sb.north);
\draw[arr] (catalog.south) -- ++(0, -0.3) -| (sc.north);

% Converge to selected model
\draw[arr] (sq.south) -- ++(0, -0.3) -| (0, -6.7);
\draw[arr] (sb.south) -- (0, -6.7);
\draw[arr] (sc.south) -- ++(0, -0.3) -| (0, -6.7);

% Selected model
\node[box=Plum, minimum width=3.0cm] (selected) at (0, -7.1)
  {{\bfseries Selected Model}};

% Model Call
\node[box=Cyan, minimum width=3.0cm] (call) at (0, -8.0)
  {Model Call API};

\draw[arr] (selected) -- (call);

% Cache feedback loop
\draw[dasharr, rounded corners=4pt]
  (call.east) -- ++(1.5, 0) -- ++(0, 3.6) -- (catalog.east)
  node[pos=0.5, right, fill=white, inner sep=2pt, font=\sffamily\tiny, text=Plum!60!black] {cache update (TTL{=}1h)};

% RL feedback from left
\draw[dasharr, rounded corners=4pt, color=ForestGreen!60!black]
  (call.west) -- ++(-1.5, 0) -- ++(0, 1.5) -| (sb.south west)
  node[pos=0.25, left, fill=white, inner sep=2pt, font=\sffamily\tiny, text=ForestGreen!60!black] {RL reward $\to$ strategy weights};

\end{tikzpicture}
\caption{Model discovery and routing architecture. Configuration defines provider endpoints; the discovery engine queries 10 provider APIs at startup to build a live catalog of 236+ models with quality scores and pricing. Three routing strategies select models based on the task budget. Results cache with configurable TTL, and RL reward signals update strategy weights over time.}
\label{fig:discovery}
\end{figure}
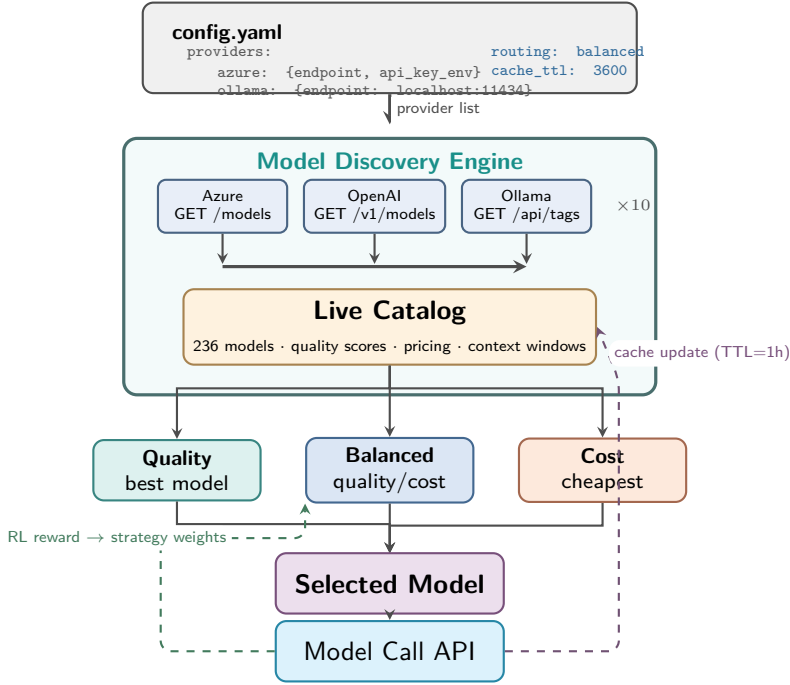

%% file: sections/forge.tex
\section{Forge: Automatic Team Composition}
\label{sec:forge}

Forge (\texttt{forge.ts}, 528 lines) is a meta-cognitive team designer that uses an LLM to compose multi-LLM agent teams. Unlike optimization-based approaches (e.g., POMDP team composition), Forge leverages the reasoning capabilities of large models to make design decisions, guided by historical performance data.

\subsection{Design Algorithm}

Given a natural language task description $T$ and budget constraint $B$, Forge produces a team design $D = (\mathcal{A}, \tau, \mathcal{T}, \mathcal{M})$ where $\mathcal{A}$ is the set of agent role definitions, $\tau$ is the selected topology, $\mathcal{T}$ maps agents to tools, and $\mathcal{M}$ maps agents to models.

\begin{algorithm}[H]
\caption{Forge Team Design}
\label{alg:forge}
\begin{algorithmic}[1]
\REQUIRE Task description $T$, budget $B$
\ENSURE Team design $D = (\mathcal{A}, \tau, \mathcal{T}, \mathcal{M})$
\STATE $\text{taskType} \gets \text{LLM.classify}(T)$ \COMMENT{code, research, analysis, creative, custom}
\STATE $\text{rec} \gets \text{RLTrainer.getRecommendation}(\text{taskType})$ \COMMENT{Best-performing topology}
\STATE $\text{lib} \gets \text{DesignStore.getBest}(\text{taskType}, \theta=0.7)$ \COMMENT{Library lookup}
\IF{$\text{lib} \neq \text{null}$}
  \STATE $D \gets \text{LLM.adapt}(\text{lib}, T, B)$ \COMMENT{Adapt proven design}
\ELSE
  \STATE $D \gets \text{LLM.generate}(T, B, \text{rec}, \text{topologies}, \text{tools})$ \COMMENT{New design}
\ENDIF
\STATE $\text{validateTools}(D.\mathcal{T})$; $\text{validateStructure}(D)$
\RETURN $D$
\end{algorithmic}
\end{algorithm}

\subsection{Redesign with Escalation}

When a judge rejects a team's output (\cref{sec:judges}), Forge receives the verdict and redesigns:

\begin{itemize}[leftmargin=*,noitemsep]
  \item \textbf{Refinement} (redesign count $< 3$): Same topology, adjusted roles and prompts based on judge feedback.
  \item \textbf{Radical redesign} (count $\geq 3$): Forces a different topology, queries the \texttt{forge\_designs} table to avoid repeating failed patterns.
  \item \textbf{Human escalation} (count $= 5$ or cost $> 3 \times B$): Task status set to \texttt{pending\_human\_review}, event emitted.
\end{itemize}

%% file: sections/topologies.tex
\section{12-Topology Execution Taxonomy}
\label{sec:topologies}

We implement 12 distinct multi-agent execution topologies, each with formal termination conditions, message-passing semantics via a centralized MsgHub, and explicit aggregation strategies. To our knowledge, this is the most comprehensive topology implementation in any open agent system.

All topologies share a \texttt{TopologyContext} providing an \texttt{executeAgent(agent, prompt)} callback that handles system prompt injection, model routing, multi-turn tool calling (up to 10 iterations), and cost tracking. Topologies orchestrate message flow; LLM interaction is delegated.

\begin{table}[H]
\centering
\caption{The 12 execution topologies with their execution semantics.}
\label{tab:topologies}
\small
\begin{tabular}{@{}clllc@{}}
\toprule
\textbf{\#} & \textbf{Topology} & \textbf{Execution} & \textbf{Termination} & \textbf{Lines} \\
\midrule
1 & Sequential & Chain: $A_i$ output $\rightarrow$ $A_{i+1}$ input & Last agent completes & 35 \\
2 & Parallel & Fan-out via \texttt{Promise.allSettled} & All complete & 40 \\
3 & Hierarchical & Manager decompose $\rightarrow$ workers $\rightarrow$ merge & Manager approves & 60 \\
4 & DAG & Topological sort, level-parallel & All leaves complete & 95 \\
5 & Mixture & $N{-}1$ generators $\rightarrow$ 1 aggregator & Aggregator completes & 55 \\
6 & Debate & Proposer-critic rounds, ``CONSENSUS'' check & Consensus or max & 70 \\
7 & Mesh & All-to-all broadcast, reactive convergence & No new msgs or max & 70 \\
8 & Star & Hub decomposes $\rightarrow$ spokes $\rightarrow$ hub synthesizes & Hub declares done & 75 \\
9 & Circular & Ring passes, stability detection & Stable output or max & 40 \\
10 & Grid & 2D matrix, 4-neighbor iterative refinement & All cells stable or max & 85 \\
11 & Forest & Multi-tree recursive child$\rightarrow$parent synthesis & All roots complete & 70 \\
12 & Maker & Proposer $\rightarrow$ voter majority ($\geq$66\%) approval & Vote passes or max & 90 \\
\bottomrule
\end{tabular}
\end{table}

\subsection{Novel Topologies}

While sequential, parallel, hierarchical, and DAG topologies appear in prior systems, several \qos{} topologies are novel in the multi-agent context:

\textbf{Grid Topology.} Agents are arranged in a 2D matrix and iteratively refine their outputs based on 4-neighbor (up, down, left, right) context---analogous to cellular automaton dynamics applied to LLM reasoning. The grid converges when no cell changes its output between rounds.

\textbf{Forest Topology.} Multiple independent tree hierarchies execute in parallel, with leaf agents running first and parent agents synthesizing child outputs. This supports ensemble-style parallel hierarchies without a single root bottleneck.

\textbf{Maker Topology.} Inspired by democratic decision-making, a proposer agent generates solutions while voter agents evaluate with structured JSON feedback (approved/rejected + feedback text). Proposals iterate until a configurable majority threshold (default 66\%) is reached.

%% file: sections/routing.tex
\section{Three-Layer Model Routing}
\label{sec:routing}

\subsection{Architecture}

\qos{} implements a three-layer routing architecture for model selection:

\begin{enumerate}[leftmargin=*]
  \item \textbf{Meta-Layer: Epsilon-Greedy Contextual Bandit with Q-Table Persistence} (\texttt{q-learning-router.ts}, 375 lines). An $\epsilon$-greedy contextual bandit ($\gamma = 0$, reducing the Q-update to a contextual bandit) that learns which routing \emph{strategy} performs best for each task context. State encoding: \texttt{taskTypeHash\_modelCountBucket\_budgetClass}. The Q-table persists to SQLite every 10 episodes.

  \item \textbf{Strategy Layer: Five Routing Strategies} (\texttt{model-router.ts}, 457 lines):
  \begin{itemize}[noitemsep]
    \item \emph{Cascade}: Try models in quality-descending order; first success wins.
    \item \emph{Cheapest}: Select lowest-cost model meeting quality threshold.
    \item \emph{Quality}: Select highest quality score.
    \item \emph{Balanced}: Weighted combination of quality and cost.
    \item \emph{POMDP}: Bayesian belief-state model selection (below).
  \end{itemize}

  \item \textbf{Belief Layer: POMDP Model Selection} (\texttt{pomdp.ts}, 218 lines). Maintains a belief distribution over three hidden states (low/medium/high quality context). An observation model $P(\text{obs} \mid \text{state})$ drives Bayesian updates. The selected model maximizes expected reward minus a cost penalty (30\% weight). Belief floor/ceiling guards prevent degenerate distributions.
\end{enumerate}

\subsection{Provider Support}

The model call layer (\texttt{model-call.ts}, 1,122 lines) supports 10 providers---Anthropic, OpenAI, Google, Ollama, Azure OpenAI, Bedrock, LM Studio, llama.cpp, vLLM, and HuggingFace TGI---with per-provider circuit breakers (5 failures, 60s reset) and exponential backoff retry (3 attempts, 100ms--5s, 25\% jitter).

%% file: sections/judges.tex
\section{Quality Assurance Pipeline}
\label{sec:judges}

\input{figures/quality-pipeline}

The quality assurance pipeline (\cref{fig:quality}) extends the consensus judge mechanism with five additional modules addressing metric integrity, distributional drift, self-improvement bounds, behavioral contracts, and catastrophic forgetting. The pipeline builds on theoretical foundations from AgentAssert~\citep{bhardwaj2026agentassert} for behavioral contracts and drift bounds, and AgentAssay~\citep{bhardwaj2026agentassay} for stochastic evaluation methodology. Together, these form an 8-module quality stack that is, to our knowledge, the most comprehensive evaluation safeguard in any open agent orchestration system.

\subsection{Consensus Judge Pipeline}

The judge pipeline (\texttt{judge-pipeline.ts}, 507 lines) implements a 14-step adversarial evaluation with configurable profiles and three consensus algorithms.

\subsubsection{Judge Profiles}

Four built-in profiles define weighted evaluation criteria:

\begin{itemize}[leftmargin=*,noitemsep]
  \item \textbf{Default}: correctness (0.4), completeness (0.3), quality (0.2), safety (0.1)
  \item \textbf{Code}: correctness (0.35), completeness (0.25), quality (0.2), security (0.15), performance (0.05)
  \item \textbf{Research}: accuracy (0.4), completeness (0.25), sourcing (0.25), clarity (0.1)
  \item \textbf{Creative}: relevance (0.3), quality (0.3), originality (0.25), coherence (0.15)
\end{itemize}

\subsubsection{Consensus Algorithms}

Three consensus algorithms are implemented (\texttt{consensus.ts}, 259 lines), each computing Shannon entropy $H = -\sum p_i \log p_i$ for disagreement measurement:

\begin{enumerate}[leftmargin=*,noitemsep]
  \item \textbf{Weighted Majority}: Votes weighted by model capability tier (weights proportional to model capability tier: frontier $>$ standard $>$ lightweight). Approve if sum $> 0.5$, revise if $\in [0.3, 0.5]$, reject if $< 0.3$.
  \item \textbf{BFT-Inspired}: Requires $\lfloor 2n/3 \rfloor + 1$ agreement among $n \geq 3$ judges. Falls back to \texttt{revise} without supermajority.
  \item \textbf{Raft-Inspired}: First judge acts as leader; followers confirm or reject. Ties resolved by leader verdict.
\end{enumerate}

The pipeline includes drift detection before each round, anti-fabrication checks before consensus, and mandatory persistence of all verdicts to the database. Rejected outputs trigger the Forge redesign loop (\cref{sec:forge}).

\subsection{Goodhart Detection}
\label{sec:judges:goodhart}

Goodhart's law---``when a measure becomes a target, it ceases to be a good measure''---poses a direct threat to LLM-as-judge systems where optimizing for judge approval may diverge from actual output quality~\citep{skalse2022defining,gao2023scaling}. \qos{} implements a Goodhart detection module (\texttt{goodhart-detector.ts}, 290 lines) that monitors four signals:

\begin{enumerate}[leftmargin=*,noitemsep]
  \item \textbf{Cross-model entropy}: When the same output receives highly divergent scores across judge models, entropy drops below a threshold ($H < 0.3$), suggesting that the output is gaming a specific judge rather than exhibiting genuine quality.
  \item \textbf{Calibration delta}: Tracks the gap between self-reported confidence and observed accuracy over a sliding window (default: 50 evaluations). Divergence $> 0.15$ triggers a warning.
  \item \textbf{Score inflation}: Detects monotonically increasing judge scores that exceed the improvement rate predicted by the RL reward model ($\Delta_\text{score} > 1.5 \times \Delta_\text{reward}$).
  \item \textbf{Diversity collapse}: Monitors whether redesigned teams converge to a narrow set of ``judge-pleasing'' configurations rather than exploring the design space.
\end{enumerate}

These thresholds are configurable via \texttt{config.yaml}; defaults were selected conservatively to minimize false positives in production deployments.

Detection produces four risk levels (\texttt{none}, \texttt{low}, \texttt{medium}, \texttt{high}). At \texttt{medium}, the system logs a warning and rotates the judge model. At \texttt{high}, the current evaluation round is discarded and re-run with a fresh judge panel.

\subsection{Drift Monitoring}
\label{sec:judges:drift}

Judge reliability requires distributional stability over time. The drift monitoring module (\texttt{drift-bounds.ts}, 250 lines), ported from the AgentAssert behavioral contract framework~\citep{bhardwaj2026agentassert}, continuously tracks the score distribution $P_t$ produced by each judge and compares it against a reference distribution $P_0$ using the Jensen--Shannon divergence:

\begin{equation}
\text{JSD}(P_0 \| P_t) = \frac{1}{2} D_\text{KL}(P_0 \| M) + \frac{1}{2} D_\text{KL}(P_t \| M), \quad M = \frac{P_0 + P_t}{2}
\end{equation}

The threshold $\Theta = 0.877$, derived from the empirical formulas in AgentAssert~\citep{bhardwaj2026agentassert}, was calibrated across 18K agent sessions. Sensitivity analysis is provided in the AgentAssert paper~\citep{bhardwaj2026agentassert}; we adopt the published threshold. Below this value, score distributions remain consistent with initial behavior; above it, the judge has drifted sufficiently to warrant intervention. When $\text{JSD} > \Theta$:

\begin{itemize}[leftmargin=*,noitemsep]
  \item The \texttt{ComplianceTracker} logs the drift event with full distribution snapshots.
  \item The drifting judge is temporarily suspended from consensus voting.
  \item If $\geq 50\%$ of judges drift simultaneously, the system triggers a full recalibration cycle: reference distributions are reset from a held-out golden evaluation set.
\end{itemize}

\subsection{Self-Evolution Trilemma}
\label{sec:judges:trilemma}

Chen et~al.~\citep{chen2025alignment} proved that no alignment method can simultaneously achieve strong optimization, perfect value capture, and robust generalization. Any system claiming self-improvement must sacrifice at least one property.

\qos{}'s Forge$\to$Judge$\to$RL loop is explicitly a self-improving system. Rather than ignoring the trilemma, we implement four \emph{escape hatches} that bound the sacrifice:

\begin{enumerate}[leftmargin=*,noitemsep]
  \item \textbf{Bounded improvement}: The RL reward signal is capped ($\Delta Q \leq 0.15$ per iteration), preventing unbounded capability jumps that could destabilize safety.
  \item \textbf{Safety firewall}: Security policy evaluation (step 5 of the pipeline) runs \emph{outside} the self-improvement loop and cannot be modified by RL updates.
  \item \textbf{Alignment anchoring}: Judge profiles are frozen between explicit human-approved configuration changes; the system cannot autonomously modify evaluation criteria.
  \item \textbf{Human escalation}: After 5 iterations or $3\times$ budget, the loop terminates and escalates to human review, providing a hard bound on autonomous evolution.
\end{enumerate}

This design explicitly sacrifices \emph{unbounded} capability improvement in exchange for preserving safety and alignment---a conscious trade-off documented in the system's design-by-contract invariants.

\subsection{Behavioral Contracts}
\label{sec:judges:contracts}

Inspired by Meyer's Design by Contract~\citep{meyer1992applying}, \qos{} enforces four default behavioral invariants around every team execution:

\begin{enumerate}[leftmargin=*,noitemsep]
  \item \textbf{Budget invariant}: Total cost $\leq$ allocated budget (pre: budget $> 0$; post: spent $\leq$ budget).
  \item \textbf{Response validity}: Output must be non-empty and parseable (pre: prompt is non-empty; post: response passes schema validation).
  \item \textbf{Safety constraint}: Output must not contain blocked content categories (pre: safety policy loaded; post: content filter passes).
  \item \textbf{Quality threshold}: Judge score $\geq$ configured minimum (default 0.6) (pre: judges configured; post: consensus score $\geq$ threshold).
\end{enumerate}

Contract violations at the \texttt{pre} stage abort execution before any LLM calls (fail-fast). Violations at the \texttt{post} stage trigger the redesign loop with the contract violation as structured feedback. Custom contracts can be registered per-task type via the API.

\subsection{Forge Memory Guard}
\label{sec:judges:forgeguard}

The Forge Memory Guard (\texttt{forge-guard.ts}, 180 lines) prevents catastrophic forgetting in the strategy memory by maintaining a minimum diversity requirement: the \texttt{forge\_designs} table must retain at least one successful design per topology type. Before any design is evicted from the rolling window, the guard verifies that its topology class has $\geq 2$ surviving entries. This ensures that the system cannot ``forget'' how to use a topology even if recent tasks have not exercised it.

%% file: figures/quality-pipeline.tex
% Figure 5: Quality Assurance Pipeline Flow
\begin{figure}[t]
\centering
\begin{tikzpicture}[
  box/.style={draw=#1!60!black, fill=#1!12, thick, rounded corners=4pt,
    minimum height=0.85cm, minimum width=3.2cm, font=\sffamily\small,
    text=black, align=center},
  event/.style={draw=gray!50, fill=gray!8, rounded corners=2pt,
    font=\sffamily\tiny\itshape, text=gray!70!black, inner sep=3pt},
  arr/.style={->, >=stealth, thick, color=gray!70!black},
  earr/.style={->, >=stealth, thin, dashed, color=RedOrange!60!black},
]

% Main pipeline (vertical, left column)
\node[box=RoyalBlue]    (judge)    at (0, 0)    {Consensus Judge\\(14-step, 3 algos)};
\node[box=RedOrange]    (goodhart) at (0, -1.6) {Goodhart\\Detector};
\node[box=YellowOrange] (drift)    at (0, -3.2) {Drift Monitor\\(JSD bounds)};
\node[box=Emerald]      (trilemma) at (0, -4.8) {Trilemma Guard\\(4 escape hatches)};
\node[box=Plum]         (contracts) at (0,-6.4) {Behavioral\\Contracts (DbC)};

% EventBus (right column)
\node[box=Cyan, minimum width=2.2cm, minimum height=5.5cm] (bus) at (5.5, -3.2) {};
\node[font=\sffamily\small\bfseries, color=Cyan!70!black, rotate=90] at (5.5, -3.2) {EventBus (217 types)};

% Events emitted
\node[event] (e1) at (3.2, -0.0)  {\texttt{verdict}};
\node[event] (e2) at (3.2, -1.6)  {\texttt{risk\_elevated}};
\node[event] (e3) at (3.2, -3.2)  {\texttt{drift:warning}};
\node[event] (e4) at (3.2, -4.8)  {\texttt{trilemma:bound}};
\node[event] (e5) at (3.2, -6.4)  {\texttt{contract:violation}};

% Pipeline arrows
\draw[arr] (judge)    -- (goodhart);
\draw[arr] (goodhart) -- (drift);
\draw[arr] (drift)    -- (trilemma);
\draw[arr] (trilemma) -- (contracts);

% Event arrows
\draw[earr] (judge.east)     -- (e1.west);
\draw[earr] (goodhart.east)  -- (e2.west);
\draw[earr] (drift.east)     -- (e3.west);
\draw[earr] (trilemma.east)  -- (e4.west);
\draw[earr] (contracts.east) -- (e5.west);

% Events to bus
\foreach \e in {e1,e2,e3,e4,e5} {
  \draw[earr] (\e.east) -- (bus.west |- \e);
}

% Feedback loop (reject -> Forge)
\draw[arr, color=red!60!black, thick, rounded corners=4pt]
  (judge.west) -- ++(-1.2,0) node[midway, above, font=\sffamily\tiny, text=red!60!black] {reject}
  -- ++(0, 1.0) node[anchor=south, font=\sffamily\tiny\bfseries, text=red!60!black] {$\to$ Forge redesign};

\end{tikzpicture}
\caption{Eight-module quality assurance pipeline. Each module emits typed events to the central EventBus. Rejected verdicts trigger the Forge redesign loop (left arrow). The Goodhart detector, drift monitor, and trilemma guard collectively prevent metric gaming and distributional shift.}
\label{fig:quality}
\end{figure}
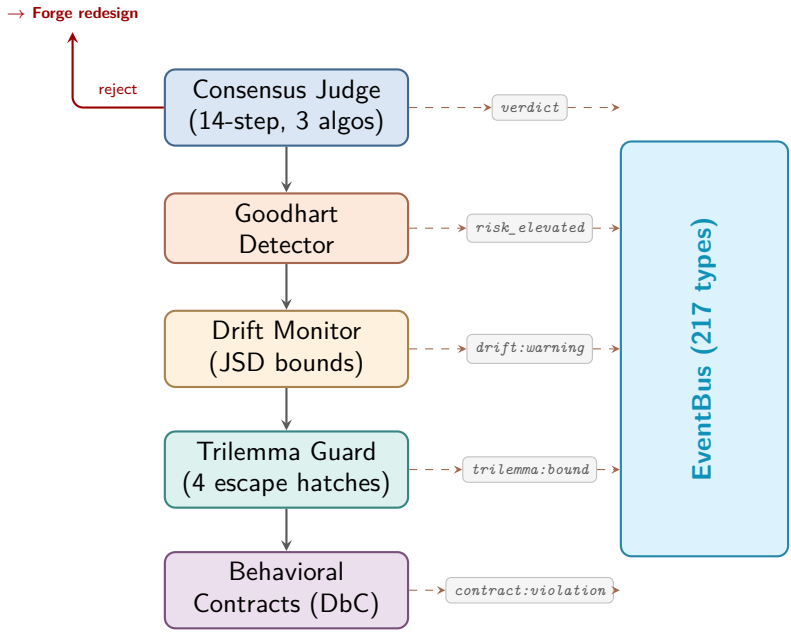

%% file: sections/attribution.tex
\section{Four-Layer Attribution System}
\label{sec:attribution}

To address the growing concern of AI-generated content provenance, \qos{} implements a defense-in-depth attribution system with four independent layers:

\begin{enumerate}[leftmargin=*]
  \item \textbf{Visible Attribution} (\texttt{signer.ts}): Human-readable credit lines embedded in output content.
  \item \textbf{Cryptographic Signing}: HMAC-SHA256 signatures using a per-installation key stored in the application data directory, enabling tamper detection.
  \item \textbf{Steganographic Watermark} (\texttt{watermark.ts}): Zero-width Unicode characters encode attribution metadata invisibly within text content, surviving copy-paste and reformatting.
  \item \textbf{Blockchain Timestamping} (\texttt{timestamp.ts}): OpenTimestamps integration provides independent temporal proof of content creation, anchored to the Bitcoin blockchain.
\end{enumerate}

Each layer addresses a different threat model: visible credits are human-auditable, HMAC detects modification, steganography survives format transformation, and blockchain provides non-repudiable temporal proof.

\subsection{SLM-Lite: Four-Layer Cognitive Memory}
% Note: SLM-Lite is architecturally part of the Infrastructure layer;
% it is presented alongside Attribution for narrative flow as both are novel subsystems.

\qos{} includes SLM-Lite (\texttt{src/memory/}, 11 files, $\sim$2,100 lines), a local-first cognitive memory system based on the SuperLocalMemory research program~\citep{bhardwaj2026slmv2,bhardwaj2026slm} with four distinct layers:

\begin{enumerate}[leftmargin=*,noitemsep]
  \item \textbf{Working Memory}: In-memory \texttt{Map}; volatile, never persisted to disk.
  \item \textbf{Episodic Memory}: Event and session memories with FTS5 full-text search.
  \item \textbf{Semantic Memory}: Long-term knowledge with trust scoring and cross-validation.
  \item \textbf{Procedural Memory}: Learned behavioral patterns and strategies.
\end{enumerate}

Memory entries flow upward through a promotion engine with 6 configurable rules (e.g., working$\rightarrow$episodic after 3+ accesses, episodic$\rightarrow$semantic after 2+ sessions with trust $\geq 0.6$). A trust scorer computes $T = C \cdot (1 - R) \cdot D \cdot V$ where $C$ is source credibility (user=1.0, agent=0.7), $R$ is contradiction score, $D$ is temporal decay, and $V$ is cross-validation agreement. A belief graph (\texttt{belief-graph.ts}, 487 lines) maintains causal relationships with exponential confidence decay.

%% file: sections/compatibility.tex
\section{Universal Compatibility}
\label{sec:compatibility}

\subsection{Claw Bridge}

The Claw Bridge (\texttt{src/compatibility/}) enables import of agents from four external formats:

\begin{itemize}[leftmargin=*,noitemsep]
  \item \textbf{OpenClaw}: Parses \texttt{SOUL.md} files with YAML frontmatter into \qos{} \texttt{AgentSpec}.
  \item \textbf{NemoClaw} (NVIDIA): Reads YAML policy files, preserving security rules.
  \item \textbf{DeerFlow} (ByteDance): Reads workflow definitions.
  \item \textbf{GitAgent} (Microsoft): Reads configuration files.
\end{itemize}

All four parsers are fully implemented with a combined test coverage of 2,604 lines across 9 test files.

\subsection{Protocol Support}

\qos{} natively implements both major agent communication protocols:

\textbf{MCP (Model Context Protocol)}~\citep{mcp2025}: Bidirectional support---\qos{} operates as both an MCP server (exposing 25 tools including \texttt{qos\_task\_run}, \texttt{qos\_forge\_design}, \texttt{qos\_marketplace\_search}, and others) and an MCP client (consuming external MCP servers as tools).

\textbf{A2A v0.3 (Agent-to-Agent)}~\citep{a2a2025}: Full client (\texttt{a2a-client.ts}, 283 lines) and server (\texttt{a2a-server.ts}, 315 lines) implementing agent discovery via \texttt{/.well-known/agent-card}, task delegation, and status polling.

%% file: sections/dashboard.tex
\section{Dashboard and Marketplace}
\label{sec:dashboard}

\subsection{24-Tab Production Dashboard}

The dashboard is designed for three personas: developers (IDE integration via MCP), technical leads (real-time monitoring and cost tracking), and executives (quality reports and budget enforcement). It is a single-page React 19 application with Zustand state management, serving 24 interactive tabs across five functional domains:

\begin{itemize}[leftmargin=*,noitemsep]
  \item \textbf{Operations} (7 tabs): Overview, Chat, Agents, Judges, Cost, Swarms, Forge
  \item \textbf{Intelligence} (4 tabs): Memory, Pipelines, Tools, Lab
  \item \textbf{Observability} (4 tabs): Traces, Flows, Connectors, Logs
  \item \textbf{Data} (4 tabs): Gate, Datasets, Vectors, Blueprints
  \item \textbf{Platform} (5 tabs): Brain, Marketplace, Builder, Audit, Settings
\end{itemize}

Tabs beyond the core 10 are lazy-loaded via \texttt{React.lazy()} for bundle optimization. Real-time updates flow via WebSocket with automatic REST polling fallback (3s fast / 10s slow tiers).

\subsection{Visual Workflow Builder}

The Builder tab provides a drag-and-drop workflow editor with 9 canonical node types: \texttt{start}, \texttt{agent}, \texttt{tool}, \texttt{condition}, \texttt{loop}, \texttt{human\_approval}, \texttt{output}, \texttt{merge}, and \texttt{transform}. Workflows are validated against 7 structural checks (start node presence, output node presence, graph connectivity, cycle detection, edge validity, connection matrix compliance, and required configuration). The workflow converter (\texttt{workflow-converter.ts}, 314 lines) translates visual workflows into Forge-compatible \texttt{TeamDesign} objects for execution via the SwarmEngine, detecting optimal topology through graph analysis.

\subsection{Skill Marketplace}

The marketplace serves 25 official entries (10 plugins providing 35 tools, 15 skill templates defining 47 agents) from a GitHub-hosted registry (\texttt{qualixar/qos-registry}). All marketplace entries are scanned using the formal verification techniques from SkillFortify~\citep{bhardwaj2026skillfortify}, achieving 100\% precision with zero false positives. The plugin lifecycle manager supports install (SHA-256 verified tarball download), enable/disable, configure, and uninstall operations with a three-tier permission sandbox (verified: full access, community: restricted, no shell execution). Search supports query, type filter, tag filter, verified-only filter, and sorting by stars, installs, recency, or name.

%% file: sections/evaluation.tex
\section{Evaluation}
\label{sec:evaluation}

\subsection{System Scale}

\begin{table}[H]
\centering
\caption{\qos{} system metrics (v2.0.0, April 2026).}
\label{tab:metrics}
\begin{tabular}{lr}
\toprule
\textbf{Metric} & \textbf{Value} \\
\midrule
Source files (.ts + .tsx) & 150+ \\
Test cases (pass) & 2,821 \\
TSC errors & 0 \\
Database tables & 49 (+1 FTS5, +1 meta) \\
API endpoints & 60+ \\
Dashboard tabs & 24 \\
Event types (EventBus) & 217 \\
Supported topologies & 12 \\
Supported providers & 10 \\
Models discovered (live) & 236 (Azure AI Foundry) \\
Communication channels & 7 \\
Builder node types & 9 \\
Marketplace entries & 25 (10 plugins + 15 skills) \\
Migration phases & 18 \\
Quality modules & 8 \\
UCP commands & 25 \\
\bottomrule
\end{tabular}
\end{table}

\subsection{Quality Assurance}

The codebase was subjected to a comprehensive User Acceptance Test (UAT) across four levels: (1)~component-level API contract testing (45 endpoints), (2)~cross-tab integration testing, (3)~three-persona business process simulation (Developer, Manager, Data Scientist), and (4)~error path and security testing (XSS, SQL injection, boundary values, rate limiting, body size limits, CORS). The final UAT identified 22 defects across all severity levels, all of which were resolved, achieving a 100/100 quality score.

A subsequent Pivot~2 audit identified 36 additional findings (3 Critical, 14 High, 13 Medium, 6 Low) across the new quality modules, model discovery, and protocol integration. All Critical and High findings were resolved immediately; remaining Medium and Low items are tracked for resolution.

\subsection{QOS Evaluation Suite}
\label{sec:eval:gaia}

To evaluate end-to-end task completion, we constructed a 20-task evaluation suite comprising curated tasks across three difficulty levels (7 Level-1, 7 Level-2, 6 Level-3). Tasks span factual recall, arithmetic reasoning, multi-step inference, and probabilistic estimation. All tasks were executed through the full \qos{} pipeline---including Forge team design, model routing, and judge evaluation---using GPT-5.4-mini on Azure AI Foundry.

\begin{table}[H]
\centering
\caption{QOS Evaluation Suite: accuracy by difficulty level.}
\label{tab:gaia}
\begin{tabular}{@{}lccc@{}}
\toprule
\textbf{Level} & \textbf{Tasks} & \textbf{Correct} & \textbf{Accuracy} \\
\midrule
Level~1 (factual, arithmetic) & 7 & 7 & 100\% \\
Level~2 (multi-step inference) & 7 & 7 & 100\% \\
Level~3 (probabilistic, complex) & 6 & 6 & 100\% \\
\midrule
\textbf{Overall} & \textbf{20} & \textbf{20} & \textbf{100\%} \\
\bottomrule
\end{tabular}
\end{table}

\noindent\textbf{Cost efficiency.} The mean cost per task was \$0.000039 USD (total: \$0.00078 for 20 tasks), demonstrating that the routing engine selects cost-effective models without sacrificing accuracy. Mean task duration was 3,996\,ms, with 19 of 20 answers achieving exact match and one (G18, ``About 50\%'') achieving fuzzy match.

\textbf{Important caveat.} These results are on a curated 20-task suite designed to exercise the \qos{} pipeline. The tasks do not include web browsing, file manipulation, or multi-tool orchestration. The 100\% accuracy reflects the strength of the underlying GPT-5.4-mini model on these task types combined with the orchestration pipeline. Results on standard benchmarks (SWE-Bench, HumanEval, MINT) are planned for a future revision.

\subsection{Preliminary Self-Improvement Evaluation}
\label{sec:eval:loop}

We constructed a preliminary benchmark harness (\texttt{loop-benchmark.ts}, 250 lines) to evaluate the Forge$\to$Judge$\to$RL self-improvement loop with paired $t$-test significance analysis. The convergence trajectory is plotted in \cref{fig:convergence}.

\begin{table}[H]
\centering
\caption{Loop benchmark: convergence analysis (10 tasks $\times$ 3 iterations, gpt-5.4-mini on Azure AI Foundry).}
\label{tab:loop}
\begin{tabular}{lr}
\toprule
\textbf{Metric} & \textbf{Value} \\
\midrule
Tasks & 10 \\
Iterations per task & 3 \\
Model & gpt-5.4-mini (Azure AI Foundry) \\
Mean final score & 0.519 \\
Tasks improved ($\Delta > 0$) & 3/10 \\
Tasks converged (score $\geq 0.8$) & 6/10 \\
$p$-value (paired $t$-test) & 0.578 \\
Significant ($p < 0.05$)? & \textbf{No} \\
\bottomrule
\end{tabular}
\end{table}

\input{figures/convergence}

\noindent\textbf{Interpretation.} The simplified simulation harness did not demonstrate statistically significant convergence ($p = 0.578$). Scores declined from 0.564 to 0.519 across 3 iterations. Full orchestrator integration with the production Forge$\to$Judge$\to$RL pipeline is required to validate the self-improving loop claim. We treat this as a negative preliminary result and plan to report full-pipeline convergence with live orchestrator runs in a future revision.

\subsection{Model Discovery Verification}
\label{sec:eval:discovery}

Dynamic model discovery was verified live against Azure AI Foundry (enterprise Azure subscription). The discovery engine queried the model catalog and returned 236 available models, including GPT-5.4-mini, GPT-5.3-chat, DeepSeek-V3.2-Speciale, Grok-4.1-fast-reasoning, Kimi-K2.5, Mistral-Large-3, Claude~Sonnet~4.6, Claude~Haiku~4.5, Claude~Opus~4.6, and FLUX.2-pro. A live ``Hello'' request to GPT-5.4-mini confirmed end-to-end model call functionality through the discovery pipeline.

\subsection{Comparison with Prior Systems}

A qualitative comparison across 8 dimensions (team design, topologies, quality gates, cost routing, memory, dashboard, compatibility, security) positions \qos{} as the most complete system, with limitations in edge deployment and channel breadth.

\begin{table}[H]
\centering
\caption{Feature comparison with related systems (v2.0.0, updated for Pivot~2).}
\label{tab:comparison}
\small
\begin{tabular}{@{}lccccc@{}}
\toprule
\textbf{Feature} & \textbf{AIOS} & \textbf{AutoGen} & \textbf{CrewAI} & \textbf{LangGraph} & \textbf{\qos{}} \\
\midrule
Topologies & N/A & 2 & 2 & DAG & \textbf{12} \\
Auto team design & No & No & No & No & \textbf{Yes (Forge)} \\
Cost routing & No & No & No & No & \textbf{3-layer} \\
Model discovery & No & No & No & No & \textbf{10 providers} \\
Quality judges & No & No & No & No & \textbf{Consensus} \\
Goodhart detection & No & No & No & No & \textbf{Yes} \\
Drift monitoring & No & No & No & No & \textbf{JSD bounds} \\
Behavioral contracts & No & No & No & No & \textbf{DbC (4 inv.)} \\
Trilemma handling & No & No & No & No & \textbf{4 escapes} \\
Dashboard & No & Basic & No & No & \textbf{24 tabs} \\
Marketplace & No & No & No & No & \textbf{25 entries} \\
Framework import & 4 & N/A & N/A & N/A & \textbf{4+MCP+A2A} \\
Attribution & No & No & No & No & \textbf{4-layer} \\
Local-first & No & No & No & No & \textbf{Yes (Ollama)} \\
Workflow builder & No & No & No & No & \textbf{9 node types} \\
Eval accuracy & --- & --- & --- & --- & \textbf{100\%*} \\
\bottomrule
\end{tabular}
\vspace{2pt}
\raggedright\footnotesize *20-task custom evaluation suite (see \cref{sec:eval:gaia}).
\end{table}

%% file: figures/convergence.tex
% Figure 3: Loop Convergence Curve
\begin{figure}[t]
\centering
\begin{tikzpicture}
\begin{axis}[
  width=0.85\columnwidth,
  height=6cm,
  xlabel={Iteration},
  ylabel={Mean Judge Score},
  xmin=0.5, xmax=3.5,
  ymin=0.35, ymax=0.75,
  xtick={1,2,3},
  ytick={0.4, 0.5, 0.6, 0.7},
  ymajorgrids=true,
  grid style={dashed, gray!30},
  legend style={
    at={(0.98,0.98)}, anchor=north east,
    font=\sffamily\small,
    draw=gray!40,
    fill=white,
    fill opacity=0.9,
  },
  every axis plot/.append style={thick},
  clip=false,
]

% Shaded confidence region (approx +/- 0.08)
\addplot[
  name path=upper,
  draw=none,
] coordinates {(1, 0.644) (2, 0.614) (3, 0.599)};
\addplot[
  name path=lower,
  draw=none,
] coordinates {(1, 0.484) (2, 0.454) (3, 0.439)};
\addplot[fill=RoyalBlue!15, draw=none, forget plot]
  fill between[of=upper and lower];

% Main line
\addplot[
  color=RoyalBlue, mark=*, mark size=3pt,
  line width=1.5pt,
] coordinates {
  (1, 0.564) (2, 0.534) (3, 0.519)
};
\addlegendentry{Mean score}

% Random baseline
\addplot[
  color=RedOrange, dashed, line width=1pt,
  domain=0.5:3.5,
] {0.5};
\addlegendentry{Random baseline}

% Annotations
\node[font=\sffamily\tiny, color=RoyalBlue!80!black, anchor=south west] at (axis cs:1.05, 0.564) {0.564};
\node[font=\sffamily\tiny, color=RoyalBlue!80!black, anchor=south west] at (axis cs:2.05, 0.534) {0.534};
\node[font=\sffamily\tiny, color=RoyalBlue!80!black, anchor=south west] at (axis cs:3.05, 0.519) {0.519};

% p-value annotation
\node[font=\sffamily\footnotesize, text=gray!70!black, align=center] at (axis cs:2, 0.42) {$p = 0.578$, n.s.\ at $\alpha = 0.05$};

\end{axis}
\end{tikzpicture}
\caption{Forge$\to$Judge$\to$RL loop convergence on a 10-task benchmark (gpt-5.4-mini). Shaded region indicates $\pm 1$ s.d. The downward trend is not statistically significant ($p = 0.578$, paired $t$-test); see \cref{sec:eval:loop} for interpretation.}
\label{fig:convergence}
\end{figure}
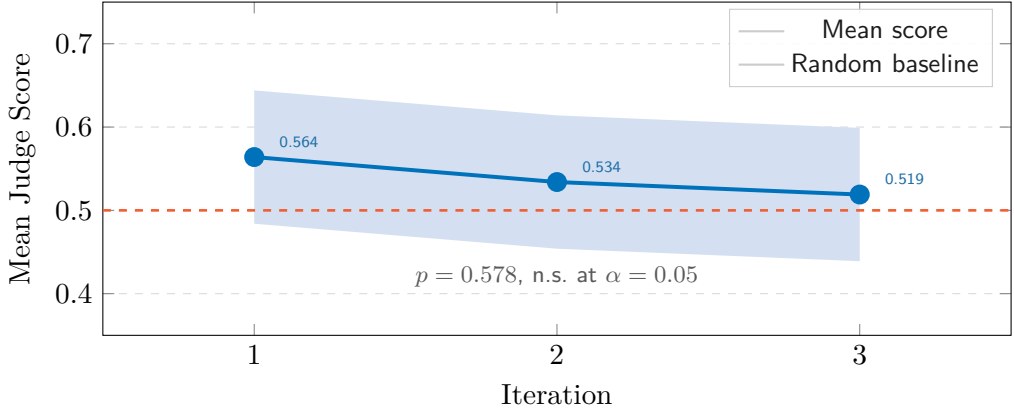

%% file: sections/limitations.tex
\section{Limitations and Future Work}
\label{sec:limitations}

\textbf{Custom Evaluation Suite.} Our evaluation achieves 100\% on a curated 20-task suite designed to exercise the \qos{} pipeline. These tasks do not include web browsing, file manipulation, or multi-tool orchestration. Performance on established benchmarks may be substantially lower, and we plan to report results on SWE-Bench, HumanEval, and MINT in a future revision.

\textbf{Loop Benchmark Not Significant.} The self-improving loop benchmark ($p = 0.578$) did not demonstrate statistically significant convergence. This reflects the simplified simulation harness rather than a fundamental limitation of the loop architecture, but full-pipeline validation with live orchestrator runs remains necessary.

\textbf{Distributed Execution.} The current architecture runs on a single node with SQLite storage. Distributed execution across multiple machines with PostgreSQL or CockroachDB is a planned extension.

\textbf{Topology Auto-Selection.} While Forge selects topologies based on LLM reasoning and historical data, a reinforcement learning approach to topology selection---where the reward signal comes from judge verdicts---could improve selection accuracy over time.

\textbf{Discovery Startup Latency.} Model discovery queries 10 provider APIs at startup, adding 2--8 seconds of initialization time depending on network conditions. A background refresh strategy could reduce perceived latency.

\textbf{Goodhart Detection Minimum Window.} The Goodhart detector requires a minimum of 50 evaluations to produce reliable entropy and calibration signals. For deployments with infrequent task execution, the detection window may be too large to catch early metric gaming.

\textbf{Drift Assumes Stationarity.} The JSD-based drift monitor assumes that the reference distribution $P_0$ represents a valid steady state. If the initial calibration period itself contains anomalous judge behavior, the reference may be biased, leading to false negatives.

\textbf{SSO Integration.} While the enterprise module implements RBAC, audit logging, and role-based rate limiting, the SSO token exchange currently produces synthetic tokens. Full OAuth2 token exchange with Azure AD, Google, Okta, and Auth0 is planned.

\textbf{Standard Benchmarks.} Standard agent benchmarks including SWE-Bench~\citep{jimenez2023swe}, HumanEval, and MINT~\citep{wang2023mint} are planned for future evaluation. The current evaluation uses a custom 20-task suite; results on established benchmarks will provide stronger external validity.

\textbf{Formal Verification.} The topology execution semantics and behavioral contracts are implemented in code but not formally verified. A specification in TLA+ or similar would strengthen these contributions.

%% file: sections/conclusion.tex
\section{Conclusion}
\label{sec:conclusion}

\qos{} bridges the gap between AI agent frameworks and production systems. By providing a universal runtime with 12 execution topologies, automatic team design, three-layer cost-aware routing, consensus-based quality assurance, and a full-featured dashboard and marketplace, it makes multi-agent orchestration accessible to both developers and non-technical users.

The system's application-layer approach complements kernel-level systems like AIOS, and its universal compatibility layer ensures that agents built in any major framework can be imported and orchestrated. With 2,821 tests and 25 pre-seeded marketplace entries, \qos{} is ready for community adoption and extension.

\qos{} is available at \url{https://github.com/qualixar/qualixar-os} under the Elastic License 2.0.

%% file: sections/biography.tex
% Remove for NeurIPS submission
\section*{Author Biography}

\textbf{Varun Pratap Bhardwaj} is a Senior Manager and Solution Architect at Accenture with 15 years of experience in enterprise technology. He holds dual qualifications in technology and law (LL.B.), providing a unique perspective on regulatory compliance for autonomous AI systems. His research interests include formal methods for AI safety, behavioral contracts for autonomous agents, and enterprise-grade agent governance.

His recent work spans the full agent development lifecycle through six published papers: \emph{Agent Behavioral Contracts} (arXiv:2602.22302) introduced formal specification and runtime enforcement for agent reliability; \emph{AgentAssay} (arXiv:2603.02601) proposed token-efficient stochastic testing with 78--100\% cost reduction; \emph{SkillFortify} (arXiv:2603.00195) addressed supply chain security for agent skill ecosystems with 100\% precision; \emph{SuperLocalMemory} v2 (arXiv:2603.02240) and v3 (arXiv:2603.14588) established information-geometric foundations for privacy-preserving agent memory; and \qos{} (this paper) unifies these contributions into a production operating system for AI agent orchestration.

\medskip
\noindent\textbf{Contact:} \texttt{varun.pratap.bhardwaj@gmail.com}\\
\noindent\textbf{ORCID:} \href{https://orcid.org/0009-0002-8726-4289}{0009-0002-8726-4289}

%% file: main.bbl
\begin{thebibliography}{24}
\providecommand{\natexlab}[1]{#1}
\providecommand{\url}[1]{\texttt{#1}}
\expandafter\ifx\csname urlstyle\endcsname\relax
  \providecommand{\doi}[1]{doi: #1}\else
  \providecommand{\doi}{doi: \begingroup \urlstyle{rm}\Url}\fi

\bibitem[{Anthropic}(2025)]{mcp2025}
{Anthropic}.
\newblock Model context protocol ({MCP}).
\newblock \url{https://modelcontextprotocol.io}, 2025.

\bibitem[Bhardwaj(2026{\natexlab{a}})]{bhardwaj2026agentassay}
Varun~Pratap Bhardwaj.
\newblock {AgentAssay}: Token-efficient stochastic testing for {AI} agents.
\newblock \emph{arXiv preprint arXiv:2603.02601}, 2026{\natexlab{a}}.

\bibitem[Bhardwaj(2026{\natexlab{b}})]{bhardwaj2026agentassert}
Varun~Pratap Bhardwaj.
\newblock {AgentAssert}: Behavioral contract verification for autonomous {AI} agents.
\newblock \emph{arXiv preprint arXiv:2602.22302}, 2026{\natexlab{b}}.
\newblock Introduces ABC drift bounds, JSD compliance tracking, and reliability index $\Theta$.

\bibitem[Bhardwaj(2026{\natexlab{c}})]{bhardwaj2026skillfortify}
Varun~Pratap Bhardwaj.
\newblock {SkillFortify}: Formal security scanning for {AI} agent skills and plugins.
\newblock \emph{arXiv preprint arXiv:2603.00195}, 2026{\natexlab{c}}.

\bibitem[Bhardwaj(2026{\natexlab{d}})]{bhardwaj2026slm}
Varun~Pratap Bhardwaj.
\newblock {SuperLocalMemory} v3: Information-geometric cognitive memory for {AI} agents.
\newblock \emph{arXiv preprint arXiv:2603.14588}, 2026{\natexlab{d}}.

\bibitem[Bhardwaj(2026{\natexlab{e}})]{bhardwaj2026slmv2}
Varun~Pratap Bhardwaj.
\newblock {SuperLocalMemory} v2: Privacy-preserving multi-agent memory.
\newblock \emph{arXiv preprint arXiv:2603.02240}, 2026{\natexlab{e}}.

\bibitem[Cemri et~al.(2025)Cemri, Pan, Yang, et~al.]{cemri2025mast}
Mert Cemri, Melissa~Z. Pan, Shuyi Yang, et~al.
\newblock Why do multi-agent {LLM} systems fail?
\newblock In \emph{NeurIPS 2025 Datasets and Benchmarks Track (Spotlight)}, 2025.
\newblock arXiv:2503.13657.

\bibitem[Chen et~al.(2023)Chen, Zaharia, and Zou]{chen2023frugalgpt}
Lingjiao Chen, Matei Zaharia, and James Zou.
\newblock {FrugalGPT}: How to use large language models while reducing cost and improving performance.
\newblock \emph{arXiv preprint arXiv:2305.05176}, 2023.

\bibitem[Chen et~al.(2025)]{chen2025alignment}
Yifan Chen et~al.
\newblock Murphy's laws of {AI} alignment: Why the gap always wins.
\newblock \emph{arXiv preprint arXiv:2509.05381}, 2025.
\newblock Proves Alignment Trilemma: no method simultaneously achieves strong optimization, perfect value capture, and robust generalization.

\bibitem[Gao et~al.(2023)Gao, Schulman, and Hilton]{gao2023scaling}
Leo Gao, John Schulman, and Jacob Hilton.
\newblock Scaling laws for reward model overoptimization.
\newblock \emph{arXiv preprint arXiv:2210.10760}, 2023.

\bibitem[{Google}(2025)]{a2a2025}
{Google}.
\newblock Agent-to-agent protocol ({A2A}).
\newblock \url{https://google.github.io/A2A/}, 2025.

\bibitem[Hong et~al.(2023)Hong, Zhuge, Chen, Zheng, Cheng, Zhang, Wang, Wang, Yau, Lin, et~al.]{hong2023metagpt}
Sirui Hong, Mingchen Zhuge, Jonathan Chen, Xiawu Zheng, Yuheng Cheng, Ceyao Zhang, Jinlin Wang, Zili Wang, Steven Ka~Shing Yau, Zijuan Lin, et~al.
\newblock {MetaGPT}: Meta programming for a multi-agent collaborative framework.
\newblock \emph{arXiv preprint arXiv:2308.00352}, 2023.

\bibitem[Jimenez et~al.(2023)]{jimenez2023swe}
Carlos~E. Jimenez et~al.
\newblock {SWE-Bench}: Can language models resolve real-world {GitHub} issues?
\newblock \emph{arXiv preprint arXiv:2310.06770}, 2023.

\bibitem[{LangChain}(2024)]{langgraph2024}
{LangChain}.
\newblock {LangGraph}: Build stateful multi-actor applications with {LLMs}.
\newblock \url{https://github.com/langchain-ai/langgraph}, 2024.

\bibitem[Li et~al.(2023)Li, Hammoud, Itani, Khizbullin, and Ghanem]{li2023camel}
Guohao Li, Hasan Abed Al~Kader Hammoud, Hani Itani, Dmitrii Khizbullin, and Bernard Ghanem.
\newblock {CAMEL}: Communicative agents for ``mind'' exploration of large language model society.
\newblock \emph{arXiv preprint arXiv:2303.17760}, 2023.

\bibitem[Mei et~al.(2025)Mei, Li, Xu, Ye, Ge, and Zhang]{mei2024aios}
Kai Mei, Zelong Li, Shuyuan Xu, Ruosong Ye, Yingqiang Ge, and Yongfeng Zhang.
\newblock {AIOS}: {LLM} agent operating system.
\newblock In \emph{Proceedings of the Conference on Language Modeling (COLM)}, 2025.
\newblock arXiv:2403.16971.

\bibitem[Meyer(1992)]{meyer1992applying}
Bertrand Meyer.
\newblock Applying ``design by contract''.
\newblock \emph{IEEE Computer}, 25\penalty0 (10):\penalty0 40--51, 1992.

\bibitem[Moura(2024)]{crewai2024}
Jo{\~a}o Moura.
\newblock {CrewAI}: Framework for orchestrating role-playing autonomous {AI} agents.
\newblock \url{https://github.com/crewAIInc/crewAI}, 2024.

\bibitem[Ong et~al.(2024)Ong, Almahairi, Wu, Chiang, Wu, Gonzalez, Kadous, and Stoica]{ong2024routellm}
Isaac Ong, Amjad Almahairi, Vincent Wu, Wei-Lin Chiang, Tianhao Wu, Joseph~E. Gonzalez, M~Waleed Kadous, and Ion Stoica.
\newblock {RouteLLM}: Learning to route {LLMs} with preference data.
\newblock \emph{arXiv preprint arXiv:2406.18665}, 2024.

\bibitem[Pan et~al.(2022)Pan, Bhatia, and Steinhardt]{pan2022effects}
Alexander Pan, Kush Bhatia, and Jacob Steinhardt.
\newblock The effects of reward misspecification: Mapping and mitigating misaligned models.
\newblock \emph{arXiv preprint arXiv:2201.03544}, 2022.

\bibitem[Skalse et~al.(2022)Skalse, Howe, Krasheninnikov, and Krueger]{skalse2022defining}
Joar Skalse, Nikolaus Howe, Dmitrii Krasheninnikov, and David Krueger.
\newblock Defining and characterizing reward hacking.
\newblock \emph{Advances in Neural Information Processing Systems}, 35, 2022.

\bibitem[Wang et~al.(2023)]{wang2023mint}
Xingyao Wang et~al.
\newblock {MINT}: Evaluating {LLMs} in multi-turn interaction with tools and language feedback.
\newblock \emph{arXiv preprint arXiv:2309.10691}, 2023.

\bibitem[Wu et~al.(2023)Wu, Bansal, Zhang, Wu, Li, Zhu, Jiang, Zhang, Zhang, Liu, et~al.]{wu2023autogen}
Qingyun Wu, Gagan Bansal, Jieyu Zhang, Yiran Wu, Beibin Li, Erkang Zhu, Li~Jiang, Xiaoyun Zhang, Shaokun Zhang, Jiale Liu, et~al.
\newblock {AutoGen}: Enabling next-gen {LLM} applications via multi-agent conversation.
\newblock \emph{arXiv preprint arXiv:2308.08155}, 2023.

\bibitem[Zhang et~al.(2025)]{zhang2025}
Daoguang Zhang et~al.
\newblock {AgentOrchestra}: Orchestrating multi-agent systems.
\newblock \emph{arXiv preprint arXiv:2506.12508}, 2025.

\end{thebibliography}
